\documentclass[10pt]{article}

\usepackage[utf8]{inputenc}
\usepackage[T1]{fontenc}
\usepackage{graphicx}
\usepackage{longtable}
\usepackage{wrapfig}
\usepackage{rotating}
\usepackage[normalem]{ulem}
\usepackage{amsmath}
\usepackage{amssymb}
\usepackage{capt-of}
\usepackage{hyperref}
\usepackage{float}
\usepackage{array}
\usepackage{tikz}
\usetikzlibrary{arrows.meta,decorations.pathreplacing,calc}
\makeatletter\newcommand{\tblscale}[1]{\begingroup\setlength{\tabcolsep}{3pt}\sbox0{#1}\resizebox{\ifdim\wd0>\linewidth \linewidth\else\wd0\fi}{!}{\usebox0}\endgroup}\makeatother

\AtBeginDocument{\setlength{\parskip}{4pt plus 1pt minus 1pt}}
\usepackage{newunicodechar}
\newunicodechar{⇒}{\ensuremath{\Rightarrow}}
\newunicodechar{↔}{\ensuremath{\leftrightarrow}}
\newunicodechar{≈}{\ensuremath{\approx}}
\usepackage{pifont}\newunicodechar{①}{\ding{172}}\newunicodechar{②}{\ding{173}}\newunicodechar{③}{\ding{174}}
\newunicodechar{④}{\ding{175}}
\usepackage{natbib}
\usepackage[margin=1in]{geometry}   
\usepackage{newtxmath}            

\usepackage{fontspec}\newfontfamily\qedfont[Path=fonts/]{DejaVuSans.ttf}\newunicodechar{≤}{\ensuremath{\leq}}\newunicodechar{≡}{\ensuremath{\equiv}}\newunicodechar{τ}{\ensuremath{\tau}}\newunicodechar{ρ}{\ensuremath{\rho}}\newunicodechar{α}{\ensuremath{\alpha}}\newunicodechar{∎}{{\qedfont\char8718}}\newunicodechar{∈}{\ensuremath{\in}}
\hypersetup{hidelinks}
\author{Zehuan Chen\thanks{Institute of Industrial Artificial Intelligence, CAS. \texttt{zhchen@iaii.ac.cn}}%
\and Chunhe Song\thanks{Institute of Industrial Artificial Intelligence, CAS. \texttt{songchunhe@sia.cn}}}
\date{October 2026}
\title{\hyphenpenalty=10000\exhyphenpenalty=10000 ORDO: Operation-level Round-aware Dynamic Ordering for MIP Presolve}
\hypersetup{
 pdfauthor={Zehuan Chen, Chunhe Song},
 pdftitle={ORDO: Operation-level Round-aware Dynamic Ordering for MIP Presolve},
 pdfkeywords={mixed-integer programming, presolve, sequence generation, learning for optimization, solver engineering, cross-domain generalization},
 pdfsubject={MIP presolve as atomic action sequence generation},
 pdfcreator={XeLaTeX},
 pdflang={English}}

\begin{document}

\maketitle
\renewcommand{\figurename}{Figure}\renewcommand{\tablename}{Table}
\renewcommand{\appendixname}{Appendix}\renewcommand{\refname}{References}\renewcommand{\bibname}{References}
\begin{abstract}

Presolve strongly affects mixed-integer programming (MIP) performance, yet learning-based methods only optimize parameter configurations and cannot express the non-commutative temporal dependencies among actions, whose default order is nearly unique on most domains, yet functionally necessary: artificially shuffling the order of the same sequence inflates the tail of the solve-time distribution by up to several-fold. We recast presolve planning as autoregressive sequence generation over a unified atomic action space, moving the decision object to action sequences; we call this framework ORDO---Operation-level Round-aware Dynamic Ordering for MIP Presolve. Its payoff is cross-domain generalization: on multiple unseen domains it attains end-to-end zero-shot speedup---to our knowledge the first for presolve action sequences---varying by domain and not explained by corpus richness, the strongest domain reaching the largest speedup once racing is added. Deployment uses sequence racing, in which candidate sequences run concurrently and the winner is kept, enabled by an execution-and-observation facility, added by modifying the SCIP source, that injects sequences along the native path and records which actions actually execute and in which round.
\end{abstract}
\section*{1 Introduction}
\label{sec:org520b4c8}

Mixed integer programming (MIP) is a core tool for combinatorial optimization. Presolve is the reduction module before branch and bound (B\&B), and its effect on overall solving performance is second only to cutting planes; our measurements show it to vary by domain (instance family)
(up to 2.4x slower than default on some domains, faster on others).
Because MIP is discrete and non-convex, presolve actions do not commute and the execution order changes the reduction outcome and the solving cost of the subsequent branch and bound.

Instance-adaptive presolve optimization has a large body of work, in four classes by decision object: manual heuristics and parameter tuning (SCIP,
\citealp{achterberg2009scip}); instance configuration tuning (ISAC, \citealp{kadioglu2010isac}; Hydra-MIP, \citealp{xu2011hydramip}; SMAC, \citealp{hutter2011smac}; L2P-MIP,
\citealp{liu2024l2pmip}); rule and subset selection (\citealp{song2024selection}); combinatorial and racing solving (\citealp{xu2011hydramip,hutter2011smac}). Order has precedents in the solving stage and on the LP side
(\citealp{tang2020cutting,zeng2025clcr}) but has not moved down to the level of individual presolve actions (atomic actions).

Within these classes, manual presets reuse one expert schedule across instances and adapt poorly to diverse instance structures \citep{achterberg2009scip,bixby2007progress}; the configuration-tuning line searches a fixed-dimensional configuration space — the closest work, L2P-MIP, formalizes presolve optimization as a 42-dimensional vector over the priority, timing and rounds of 14 routines \citep{liu2024l2pmip}, an output that carries no action call order and cannot intervene in the call order of constraint handler callbacks; rule and subset selection decides only which routines participate, not in what order \citep{song2024selection}; and combinatorial/racing methods build their candidates at the level of configurations or complete solving strategies, truncating the heavy-tailed runtime of a single policy with concurrent candidates but never descending to the presolve action-sequence level \citep{xu2011hydramip,hutter2011smac}.
The common limitation of the four classes is that none can express the temporal dependencies among actions: all of them stop at the configuration-vector or participation-subset level, so they capture no commonality across problem classes and cannot generalize; our candidate is a presolve action sequence (\S2.5; its independent value in \S3.4). A full review with the item-by-item comparison is in Appendix B.10.

This paper therefore reformulates MIP presolve planning from parameter configuration optimization into conditional autoregressive sequence generation over a unified atomic action space: the decision object is an action sequence
satisfying the execution contract, and deployment uses sequence racing, which executes multiple candidate sequences concurrently and keeps the winner, to truncate the tail risk of single-path prediction.

The main contributions of this paper are as follows.

\begin{enumerate}
\item Zero-shot generalization from sequence generation. We recast presolve optimization from a search over configuration parameters to conditional autoregressive generation over a unified atomic action space, so the model generates the presolve action sequence directly from instance features rather than selecting a configuration; a zero-shot evaluation on 6 leave-one-out domains and 1 naturally unseen domain shows end-to-end speedup on multiple unseen domains, to our knowledge the first such result for presolve action sequences.
\item From single-path prediction to sequence racing. Prior racing work builds its candidates only at the level of configurations or complete solving strategies and never descends to the presolve action-sequence level; we move racing down to that level and decompose its gain into candidate quality and competition amplification.
\item An execution-and-observation facility for presolve sequences. Black-box solvers expose no interface for injecting an action sequence, so we modify the SCIP 10.0.1 source to inject a given sequence along the native path and to record which actions actually execute and in which round.
\end{enumerate}

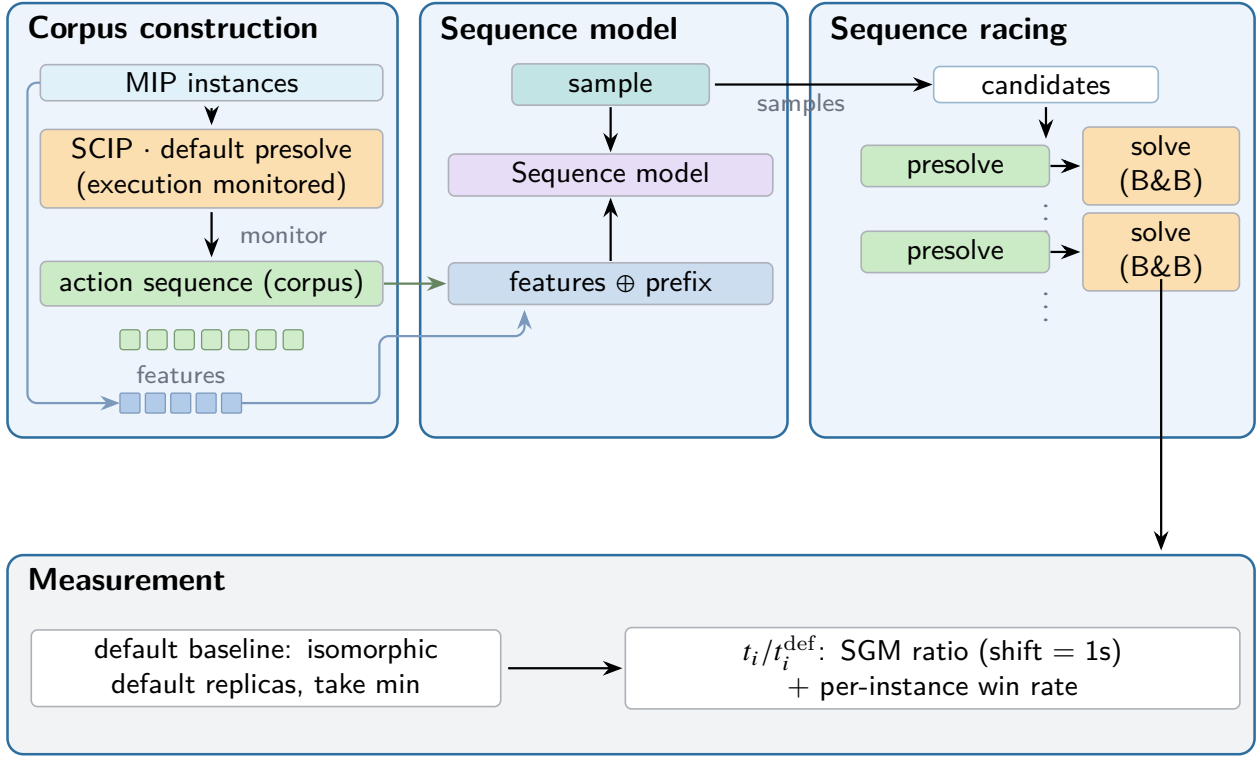
\begin{figure}[!t]
\centering\resizebox{\linewidth}{!}{\begin{tikzpicture}[font=\sffamily\small, >=Stealth, every node/.style={inner sep=1pt}]
\definecolor{cPanel}{HTML}{EDF4FB}
\definecolor{cBand}{HTML}{F5F7F9}
\definecolor{cFrame}{HTML}{2E6E9E}
\definecolor{cData}{HTML}{CBE7F5}
\definecolor{cFeat}{HTML}{8FB4DE}
\definecolor{cSolver}{HTML}{F8C97C}
\definecolor{cSeq}{HTML}{A9DCA0}
\definecolor{cModel}{HTML}{D8C7F0}
\definecolor{cSample}{HTML}{A8D8D8}
\definecolor{cMeas}{HTML}{F1F3F5}
\definecolor{cNote}{HTML}{6B7280}

\tikzset{
  seg/.style  ={draw=cFrame, fill=cPanel, rounded corners=5pt, line width=0.8pt},
  box/.style  ={draw=cNote!55, rounded corners=2pt, line width=0.5pt, align=center,
                font=\sffamily\small, inner sep=2.5pt},
  ttl/.style  ={anchor=north west, font=\sffamily\bfseries\normalsize},
  flow/.style ={->, line width=0.7pt, rounded corners=3pt},
  lab/.style  ={font=\sffamily\footnotesize, align=center, text=cNote},
}

\draw[seg] (0.10,3.60) rectangle (4.40,8.40);
\draw[seg] (4.65,3.60) rectangle (8.70,8.40);
\draw[seg] (8.95,3.60) rectangle (13.85,8.40);
\node[ttl] at (0.28,8.30) {Corpus construction};
\node[ttl] at (4.83,8.30) {Sequence model};
\node[ttl] at (9.13,8.30) {Sequence racing};

\node[box, fill=cData!60, text width=3.6cm] (inst) at (2.35,7.52) {MIP instances};
\draw[flow] (2.35,7.24) -- (2.35,7.05);
\node[box, fill=cSolver!60, text width=3.6cm] (solver) at (2.35,6.57)
      {SCIP $\cdot$ default presolve\\ (execution monitored)};
\draw[flow] (2.35,6.09) -- (2.35,5.59);
\node[lab, anchor=west] at (2.62,5.84) {monitor};
\node[box, fill=cSeq!60, text width=3.6cm] (corpus) at (2.35,5.30)
      {action sequence (corpus)};
\foreach \i in {0,...,6}
  \filldraw[draw=cSeq!70!black, fill=cSeq!55, rounded corners=1pt, line width=0.4pt]
           (1.34+0.30*\i,4.57) rectangle (1.56+0.30*\i,4.79);
\foreach \i in {0,...,4}
  \filldraw[draw=cFeat!85!black, fill=cFeat!70, rounded corners=0.5pt, line width=0.4pt]
           (1.34+0.28*\i,3.87) rectangle (1.56+0.28*\i,4.09);
\draw[flow, draw=cFeat!85!black] (0.46,7.52) -- (0.32,7.52) -- (0.32,3.98) -- (1.34,3.98);
\node[lab, anchor=south] at (2.01,4.16) {features};
\draw[flow, draw=cSeq!60!black] (4.25,5.30) -- (4.94,5.30);
\draw[flow, draw=cFeat!85!black] (2.68,3.98) -- (4.20,3.98) -- (4.20,4.72) -- (5.80,4.72) -- (5.80,5.01);

\node[box, fill=cSample!70, text width=2.0cm] (sample) at (6.75,7.50) {sample};
\draw[flow] (6.75,7.25) -- (6.75,6.75);
\node[box, fill=cModel!60, text width=3.4cm] (model) at (6.75,6.50) {Sequence model};
\node[box, fill=cFeat!45, text width=3.4cm] (feat) at (6.75,5.30)
      {features $\oplus$ prefix};
\draw[flow] (6.75,5.55) -- (6.75,6.25);
\draw[flow] (7.90,7.50) -- (10.30,7.50);
\node[lab, anchor=north] at (8.85,7.44) {samples};

\node[box, fill=white, draw=cFrame!50, text width=2.3cm] (hub) at (11.55,7.50)
      {candidates};
\draw[flow] (11.55,7.25) -- (11.55,6.90);
\node[box, fill=cSeq!60, text width=1.9cm] (c1) at (10.55,6.60) {presolve};
\draw[flow] (11.60,6.60) -- (11.93,6.60);
\node[box, fill=cSolver!60, text width=1.55cm] (s1) at (12.82,6.60) {solve (B\&B)};
\node[lab] at (11.55,6.12) {$\vdots$};
\node[box, fill=cSeq!60, text width=1.9cm] (c2) at (10.55,5.65) {presolve};
\draw[flow] (11.60,5.65) -- (11.93,5.65);
\node[box, fill=cSolver!60, text width=1.55cm] (s2) at (12.82,5.65) {solve (B\&B)};
\node[lab] at (11.55,5.15) {$\vdots$};

\draw[seg, fill=cMeas] (0.10,0.10) rectangle (13.85,2.30);
\node[ttl] at (0.28,2.20) {Measurement};
\draw[flow] (12.82,5.35) -- (12.82,2.35);
\node[box, fill=white, text width=5.0cm] (base) at (2.95,1.05)
      {default baseline: isomorphic default replicas, take min};
\draw[flow] (5.62,1.05) -- (6.88,1.05);
\node[box, fill=white, text width=6.6cm] (sgm) at (10.30,1.05)
      {$t_i/t_i^{\mathrm{def}}$: SGM ratio (shift = 1s)\\
       + per-instance win rate};

\end{tikzpicture}%
}
\caption{The four-stage pipeline of the method in this paper. Corpus construction and sequence modelling are each performed once offline; the online part performs only per-instance decoding and sequence racing; the measurement layer takes the solver-internal time of the isomorphic default strategy as the denominator and reports the SGM ratio and the per-instance win rate.}
\label{fig:pipeline}
\end{figure}
\section*{2 Method}
\label{sec:org4c99265}

We refer to the framework of this paper as ORDO (Operation-level Round-aware Dynamic Ordering for MIP Presolve), after the Latin \emph{ordo}: the decision object is the order of presolve actions, and what races at deployment time is a generated order. Figure \ref{fig:pipeline} shows the four-stage pipeline of the method.

\subsection*{2.1 Problem Definition and Formalization}
\label{sec:org6928067}

Property 1 (Non-commutativity of actions). Let \(f\) and \(g\) be two presolve atomic actions and \(x\) an instance. Unless the constraint/variable sets they act on are completely disjoint,
in general \(f(g(x)) \neq g(f(x))\), with different reduction effects and relaxation strengths. The cause is the discrete non-convexity of MIP:
one action's result changes another's applicability conditions and strength.
Presolve planning therefore cannot be characterized by the set or frequency of its actions; the temporal topology must be retained as an independent variable.

All notation and the meaning of every formula used in this paper are collected in Appendix C.
\subsection*{2.2 Model Architecture}
\label{sec:orgd828e6f}
\begin{figure}[!t]
\centering\resizebox{\linewidth}{!}{\input{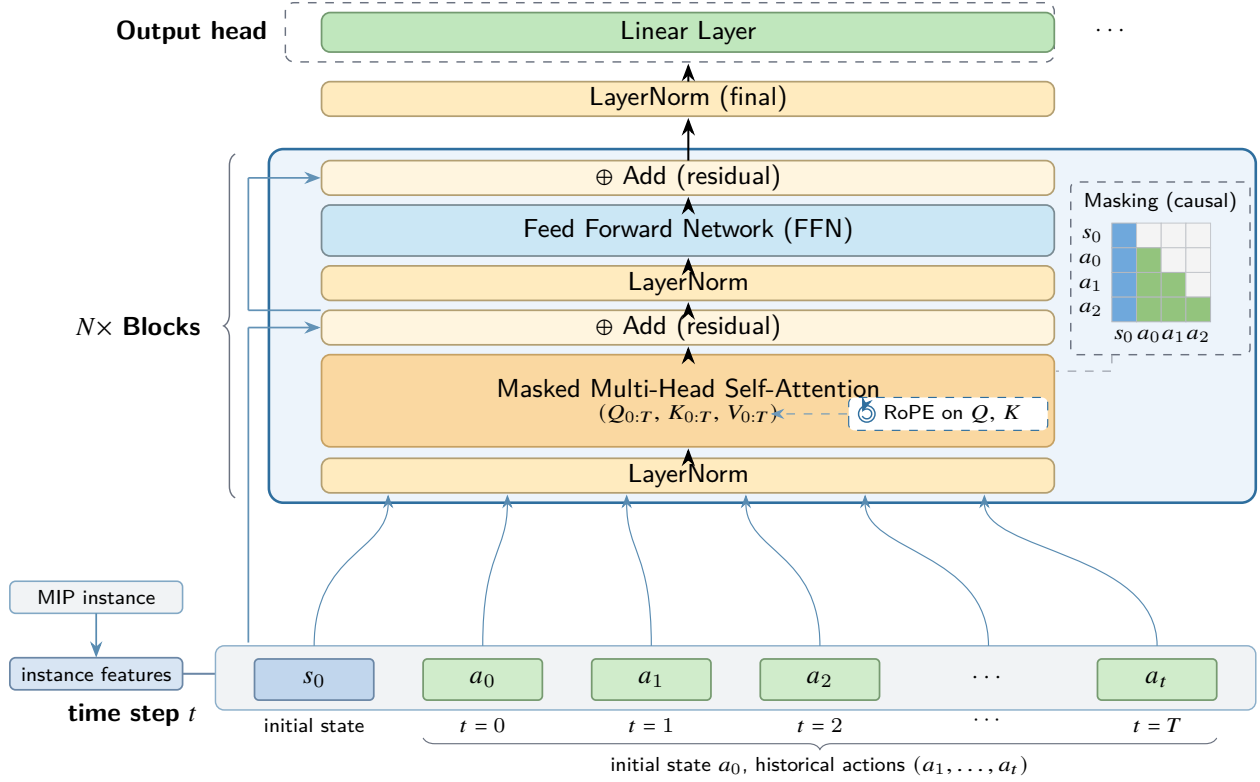}}
\caption{Structure and conditioning of the sequence generation model. Instance features are fed as the initial state concatenated with the historical action tokens, and causal self-attention (rotary position embedding, RoPE, applied to \(Q/K\)) autoregressively predicts the next item; round boundaries and the terminator share one output space with the actions, so that "in what order and at which round to execute" is given by the same distribution.}
\label{fig:architecture}
\end{figure}

Problem (Generation of presolve execution sequences). Given an instance description \(x\), seek a generative distribution \(\pi(\cdot \mid x)\) whose samples
\(\tau \sim \pi(\cdot \mid x)\) minimize the total solving time of that instance.

We reformulate this generation from a parameter optimization into a sequence modelling problem: the decision object is the complete action sequence of Definition 1, and the learning objective is autoregressive conditional generation
\begin{equation}\label{eq:cond-gen}
p_\theta(a_t \mid a_{0,\ldots,t-1}, x)
\end{equation}
This brings two structural differences against existing formulations: the decoupling of representational capacity and the expansion of the decision domain; both are developed in Appendix A.4. Figure \ref{fig:architecture} shows the structure and the conditioning of the model.

Training follows standard next-token prediction: the parameters are fitted by the cross-entropy loss of autoregressive language modelling over action sequences recorded from the solver, conditioned on the instance features; the corpus comes from executing the default strategy and keeping the actions that take effect, so it characterizes the behaviour of the default policy rather than a distribution over optimal sequences (Appendix B.11, B.12.3).
\subsection*{2.3 Unified Atomic Action Space}
\label{sec:org1e0b246}

Definition 1 (Presolve action space). Let the solver expose three families of intervenable plug-ins at the presolve stage:
\begin{equation}\label{eq:action-space}
\mathcal{A} = \mathcal{A}_{\mathrm{presol}} \cup \mathcal{A}_{\mathrm{prop}} \cup \mathcal{A}_{\mathrm{cons}}
\end{equation}
where \(\mathcal{A}_{\mathrm{presol}}\) are the independent reduction routines, \(\mathcal{A}_{\mathrm{prop}}\) the domain propagators and \(\mathcal{A}_{\mathrm{cons}}\) the constraint reduction/propagation callbacks of the constraint handler, the SCIP components that enforce constraint classes and tighten their domains. Note: the facility intercepts and executes all three families at the interface level;
our corpus and replay path use two sources, independent routines and constraint handler callbacks; propagators yield no recorded action inside the pure presolve window, and the replay path disables probing-type trial and error (Appendix B.12).
A presolve plan is a sequence \(\tau = (a_1, \dots, a_T),\ a_t \in \mathcal{A} \cup \{\bot_{\mathrm{round}}\} \cup \{\bot_{\mathrm{eos}}\}\),
where \(\bot_{\mathrm{round}}\) marks a round boundary and \(\bot_{\mathrm{eos}}\) termination.

Remark. The configuration space cannot intervene in the call order of constraint handler callbacks, so the decision object here is the action sequence; we make no claim that configuration schemes perform worse (Appendix A.4).

Representational corollary. Behaviour cloning on the sequence outputs action tokens whose vocabulary (presolver routine callbacks / constraint handler callbacks / round boundaries) is shared across domains, so the model reuses one action syntax and round structure without redefining the decision object on a new domain. A configuration-space learner instead outputs a parameter vector set by each routine's priority / timing and the round cap: it carries no execution order and shares no vocabulary across domains, so what is learnable stops at choosing which parameter set to use for an instance, not the ordering structure of when and at which round to execute.

The corpus bounds this capability: it comes from the default strategy's own trajectory, whose round structure is nearly unique, so the corpus contains no example of a round structure beyond that policy's, and gains from adding rounds are not measured here; the round-boundary token itself is part of the action space (Appendix B.12.2). Two independent measurements agree (§3.3, §3.5), so the carrier of the gain is the scheduling structure itself.

Why imitation rather than reinforcement learning: the corpus is precisely the default strategy's own execution trajectory, so behaviour cloning is the natural landing point for a decision object that did not previously exist, whereas reward-driven learning would need a repeatable reward and a stable policy space while every rollout here costs a full solve; optimizing the sequence model by reward is left to future work.
\subsection*{2.4 Sequence Execution Facility for Presolve Sequences}
\label{sec:org666d949}

Definition 2 (Execution contract). Let \(E_c(\tau, x)\) denote the execution mapping of sequence \(\tau\) on instance \(x\) and \(c\) the contract: \(E_0\) executes only \(\tau\); \(E_D\) executes \(\tau\) followed by
the default presolve as fallback. The execution interface of sequences is in Appendix B.11.2; it was added by modifying the SCIP 10.0.1 source at the three plug-in execution entry points. SCIP is used because it is, among mainstream solvers, the one with modifiable source and a mature presolver ecosystem at plug-in granularity; closed-source solvers offer no such hook. The facility is model-agnostic: hooks at the three plug-in execution entry points plus six exported C interfaces exposed through the PySCIPOpt binding let any sequence over this action space be injected and recorded externally (to be released in a follow-up public version).
\subsection*{2.5 Deploying Presolve Sequence Racing}
\label{sec:org797cea8}

By the execution contract of §2.4, any number of different sequences can be executed on the same instance, so "multi-route" has a definite meaning at the presolve level: one route is one candidate to be executed — an action sequence, or the sequence followed by the default presolve as fallback — and routes differ only in the sequence and the execution contract, not in solver configuration. With the actual completion time as the arbitration quantity, racing takes
\(\arg\min_{r \in \mathcal{R}} T(r)\) over the candidate set \(\mathcal{R}\).

Our concrete implementation runs \(2K+1\) paths: one fully-presolve-disabled arm (np) + \(K\) custom-sequence arms (executing the model's sequence only) + \(K\) default-fallback arms (the sequence followed by the default presolve). Two design points: the np arm puts the alternative explanation "doing less or no presolve" inside the pool (its control is in §3.4 and Appendix B.5), and the fallback arms keep candidates in which the sequence is only a lead-in after which the default presolve still runs. The pool contains no standalone default strategy, so racing provides no structural safety net: correctness is borne by the semantic consistency of sequence execution, not by
racing.

The deployment form and the measurement protocol are shown in Figure \ref{fig:racing} and formalized in Appendix A.1. \(K\) is an engineering choice; we make no claim about its optimality, since its compression effect arises only at fixed total compute (Remark 3 of Appendix B.5). We claim no novelty of the racing mechanism itself: what we claim is the candidate object (presolve action sequences) and its deployment and measurement protocol.

\begin{figure}[!ht]
\begin{center}
\includegraphics[width=\linewidth]{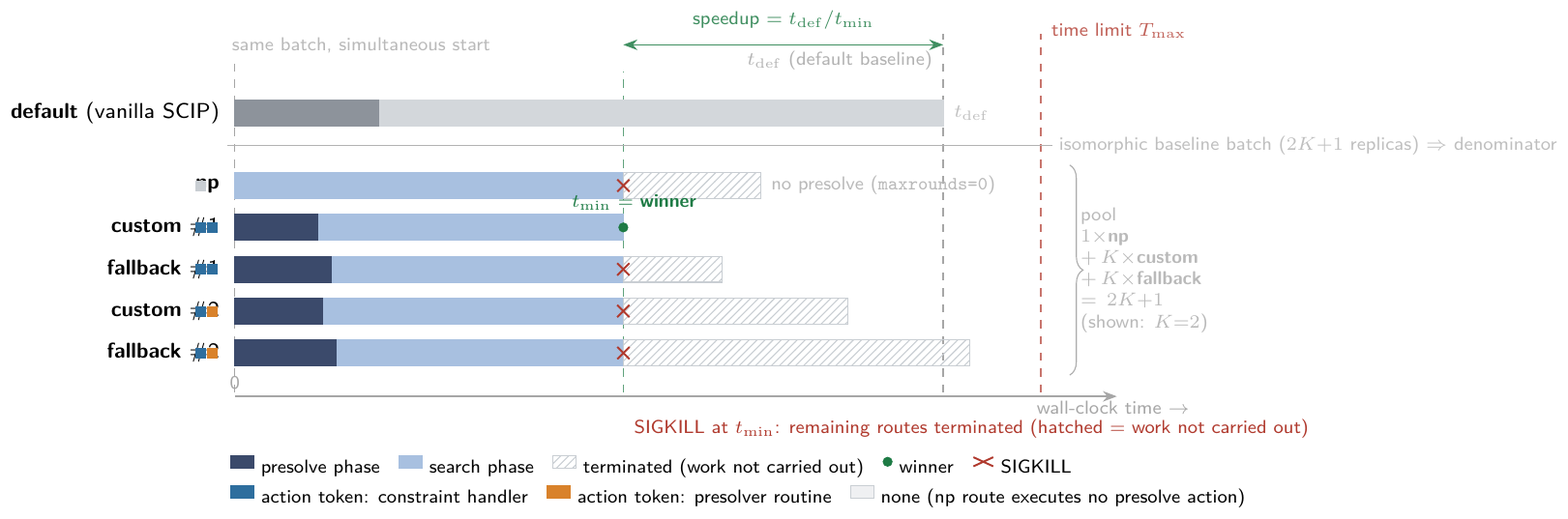}
\end{center}
\caption{The deployment form and measurement protocol of concurrent racing. Pool = np + content×\(K\) + fallback×\(K\) = \(2K+1\) paths launched together; the first to finish wins and the others are withdrawn (SIGKILL); the numerator is the solver-internal time of the winner, and the denominator is the internal clock of the replica of the isomorphic default strategy selected as the first to finish, so the speedup shown in the figure is the ratio reading with which that instance enters SGM.}
\label{fig:racing}
\end{figure}

\section*{3 Experiments and Results}
\label{sec:org2b5c4b7}
\subsection*{3.1 Experimental Setup}
\label{sec:orgf148815}

Dataset and instance sets: experiments run on SCIP 10.0.1 and use instances from the Distributional MIPLIB (D-MIPLIB) dataset of \citep{huang2024distributional}, a benchmark of domain-wise instance families. Seven MIP domains are used: CA, GISP, CFLP, MIS, MVC, SC (100 instances each) and NNV (550), all from the easy val split.

Evaluation metrics: the main metric is the shifted geometric mean (SGM) ratio against the default strategy, \(SGM(t)=\left(\prod_i (t_i+1)\right)^{1/n}-1\) (shift=1s, per instance, timeouts counted at that instance's time limit), a ratio below 1 being a speedup; we also report the per-instance win rate, the fraction of instances faster than default.

Online comparison follows one uniform protocol: both sides use 5 online cores and the same time limit, the solver-internal clock is used throughout, and the denominator is always the first finisher's clock among the same-batch 5 isomorphic default replicas of the default strategy on the same instance and settings, replicated independently, so absolute values must not be mixed across arms or batches. The only asymmetry lies in the numerator: an opponent arm copies one configuration 5 times and takes its fastest finisher (a replica-best-of bonus), whereas our racing winner contributes a single reading and is not re-minimised; the single-path side likewise reads the best of 5 replicas of the same content (§3.3, Appendix B.5). This asymmetry pushes our own reading down and we do not turn it to our advantage. The full protocol and the measured size of that bonus are in Appendix B.2.2.
Runtime environment: all readings were collected on a single cloud instance, an ecs.c9i.16xlarge with 64 vCPU and 128 GB RAM, with all experiments executed inside a custom Docker image based on Arch Linux on a host running Ubuntu 24.04, with a self-compiled SCIP 10.0.1 (source-modified for sequence execution). The sequence model is a 1.6M-parameter lightweight decoder-only network, generated online by CPU forward inference with stochastic decoding at temperature 1.0, at a per-instance inference overhead below 0.1\% of the total time. A necessary condition for replication: the runs must be performed on a machine with fair CPU allocation and no heterogeneous big.LITTLE cores, otherwise the timing readings are contaminated and the speedups reported here are not reproducible; on heterogeneous CPUs, explicit core pinning is required and the cores behind the concurrent paths must be equivalent. The full environment list and the inference-overhead breakdown are in Appendix B.2.1.

\subsection*{3.2 Correctness Verification}
\label{sec:org2782b25}

The sequence injection changes neither the solving semantics nor the optimal solution: 658/658 instances are bitwise identical to vanilla, the per-replica objective-value difference is identically 0 over 1500 solves, and there are 0 inconsistencies (Appendix B.1).
\subsection*{3.3 Empirical Evidence for the Duality of Order}
\label{sec:org87b81e9}

Finding (the duality of order): the default trajectory is nearly order-unique, yet that order is functionally necessary. In most domains such as SC / GISP / CFLP the instance-level sequence space is nearly zero-dimensional (effective number of sequences \(N_{\mathrm{eff}}\) ≈ 1), while CA / MMCN / MIRP show instance-level diversity (\(N_{\mathrm{eff}}\) up to 800); the order-invariance degree \(\rho_{\mathrm{order}}\) falls in [1.00, 1.25] across all domains. Necessity: artificially shuffling the full token order of the same sequence degrades all five domains with heavy tails (\(p_{90}\) ratio 1.22--2.86, \(p_{99}\) 1.43--5.15, max 6.8; Appendix B.3), so near-uniqueness and functional necessity are not contradictory but the two halves of the duality, which explains why this dimension has long lacked a learnable signal. Per-domain catalogue, quantiles and the ordered/shuffled readings (Table \ref{tab:order-ablation}) are in Appendix B.3.
\subsection*{3.4 Decomposing the Gains of the Racing Mechanism}
\label{sec:org888482c}

Appendix B.5 (Table \ref{tab:k-ablation}) gives the gain decomposition and the pure sequence execution experiment: the gain comes from added candidates rather than swapping contents, and pure single-path execution beats the default on all five domains (largest margin NNV 0.8377) with racing adding \textasciitilde{}8\% on top.

The applicability condition of racing is given in Appendix B.7: the gain is a tail effect, whose direction is fixed by the min-of-\(K\) calibration of Proposition 1, so racing is worth deploying only when the target domain's default strategy carries material tail risk; MVC and CFLP are the counter-examples in our own readings.

\subsection*{3.5 Zero-Shot Cross-Domain Transfer}
\label{sec:org93b9945}

Across the 7 cross-domain points (6 leave-one-out constructions plus 1 naturally unseen domain, NNV) the verdict is 4 accelerated, 2 tied and 1 failed, all measured on the same batch with the internal clock and the limits aligned on both sides. The strongest point is the naturally unseen domain, whereas the domain with the richest training corpus (CA) fails, so every point is reported individually rather than as an average (Appendix B.12.4).

NNV (training set entirely 0-1 variables, target 96\% continuous) attains SGM=0.7181 and a per-instance win rate of 73.6\% (405/550) over the 550 keys of the protocol split in Appendix A.2 and B.4, with limits aligned on both sides.

\begin{table}[!htb]
\caption{Generalization test and results (training-domain composition in Table \ref{tab:model-corpus}); per-instance win rates 28\%--95\% over the 6 leave-one-out constructions and over all 7 points.}
\label{tab:generalization}
\centering
\small
\tblscale{%
\begin{tabular}{llrrr}
\hline
\shortstack{training\\set} & \shortstack{test\\(target)} & SGM & \shortstack{win rate\\vs default} & \shortstack{sign-test\\\(p\)} \\
\hline
all 8 & CA (in-domain) & 0.9804 & 55\% (55) & 0.3682 \\
all 8 & GISP (in-domain) & 1.0050 & 44\% (44) & 0.2713 \\
all 8 & NNV (zero-shot) & 0.7181 & 73.6\% (405) & 2.0e-29 \\
8 \(\setminus\) \{GISP\}\(^{\ast}\) & GISP (LOO) & 0.9357 & 70\% (70) & 7.9e-05 \\
8 \(\setminus\) \{CA\}\(^{\ast}\) & CA (LOO) & 1.1121 & 28\% (28) & 1.3e-05 \\
8 \(\setminus\) \{CFLP\} & CFLP (LOO) & 0.8065 & 82\% (82) & 6.1e-11 \\
8 \(\setminus\) \{MIS\} & MIS (LOO) & 1.0140 & 39\% (39) & 0.0352 \\
8 \(\setminus\) \{MVC\} & MVC (LOO) & 0.9964 & 47\% (47) & 0.6173 \\
8 \(\setminus\) \{SC\}\(^{\ast}\) & SC (LOO) & 0.7725 & 95\% (95) & 1.3e-22 \\
\hline
\end{tabular}}
\end{table}

Table note: \(n\) = 100 instances per domain, except NNV (550); win-rate cells give the percentage and the count; the two-sided sign test (Appendix B.4) gives the verdict (\(p<0.05\) with SGM \(>1\) a loss, SGM \(<1\) a speedup, \(p\ge 0.05\) a tie), and the SGM deviation gives its direction and magnitude.

All rows share identical model structure and hyperparameters, differing only in the training-domain set (Table \ref{tab:model-corpus}). "All 8" is the full training set (CA, GISP, CFLP, MIS, MVC, SC, MMCN and LB; MMCN / LB are trained on but never evaluated), and "8 \(\setminus\) \{X\}" is that set minus X. The 3 rows marked \(^{\ast}\) are the SC / CA / GISP leave-one-out models: the row of this paper in Table \ref{tab:opponents-config} is per domain that domain's leave-one-out zero-shot reading, so each domain uses its own leave-one-out model rather than a single model. The counterexample and tie-point readings, the per-protocol cross-domain readings and the limit-alignment protocol are in Appendix B.4, where the \(p\) column (a two-sided sign test) is defined.
\subsection*{3.6 Comparison Against Prior Work}
\label{sec:org78fa92f}

\begin{table}[!htb]
\caption{Reimplementations of the setups of prior work (configuration-space class; SC / CA / GISP, 100 val instances; solver-internal time; each arm's numerator is the internal clock of the fastest of 5 same-configuration replicas launched concurrently in the same-batch evaluation, while the racing arm of this paper takes the winning pool path's single reading — protocol details in Appendix B.8; values = SGM relative to the same-batch default strategy, win rates in parentheses)}
\label{tab:opponents-config}
\centering
\small
\tblscale{%
\begin{tabular}{>{\raggedright\arraybackslash}p{4.0cm}>{\raggedright\arraybackslash}p{3.2cm}rrr}
\hline
Opponent &\shortstack{Category\\and citation}& SC & CA & GISP \\
\hline
\multicolumn{5}{l}{\emph{Prior work: reimplementations of the setups in the configuration space}} \\
L2P-MIP-style domain-level configuration & L2P-MIP \citeyearpar{liu2024l2pmip} & 0.984 (62\%) & 0.985 (57\%) & 0.985 (45\%) \\
L2P-MIP-style per-instance configuration & L2P-MIP \citeyearpar{liu2024l2pmip} & 0.857 (92\%) & 0.875 (93\%) & 0.787 (100\%) \\
SMAC-style configuration search & SMAC \citeyearpar{hutter2011smac} & 0.977 (62\%) & 1.018 (52\%) & 0.972 (60\%) \\
Song \& Gu-style subset selection & Song \& Gu \citeyearpar{song2024selection} & 0.993 (60\%) & 0.998 (54\%) & 0.979 (54\%) \\
ISAC-style per-class configuration & ISAC \citeyearpar{kadioglu2010isac} & 0.987 (57\%) & 1.005 (59\%) & 0.969 (55\%) \\
Hydra-MIP-style configuration selection & Hydra-MIP \citeyearpar{xu2011hydramip} & 0.989 (53\%) & 1.131 (29\%) & 0.976 (53\%) \\
\hline
\multicolumn{5}{l}{\emph{Method of this paper}} \\
sequence generation + racing & --- & 0.773 (93\%) & 1.102 (30\%) & 0.921 (67\%) \\
\hline
\end{tabular}}
\end{table}

This section uses one batch of instances, one denominator and one concurrent load. The configuration-space opponents are in Table \ref{tab:opponents-config} and the racing-deployment opponents in Table \ref{tab:opponents-deploy} (the latter in Appendix B.8). The comparison's domain-selection rule, protocol boundaries and prior work's reading status are in Appendix B.8; one protocol asymmetry must be read alongside: the row of this paper is a leave-one-out zero-shot reading (per domain: the 3 rows marked \(\ast\) in Table \ref{tab:generalization}), whereas every opponent arm searches configurations and trains its selector on that domain's train split, where the per-instance arm uses 50 train instances with 200 evaluations each. This difference is unfavourable to this paper and we do not interpret it in our favour.

Both constructions were measured: the same table reports the in-domain rows (all 8: CA 0.9804 / GISP 1.0050) alongside the leave-one-out rows, and the relative direction is domain-dependent — on CA the in-domain reading is the better of the two, on GISP the leave-one-out reading is — so the choice of construction is not uniformly favourable or unfavourable to this paper. This table therefore serves positioning, not a direct win/loss reading: our zero-shot method beats the entire configuration-class range on SC (0.773 versus 0.857--0.993), while gaps remain on GISP / CA (0.921 versus 0.787--0.985; 1.102 versus 0.875--1.131). The gaps reflect a difference in setting — "with" vs. "without target-domain corpus" — not a method comparison under one identical problem definition.

The gains cannot be replaced by simpler means. (i) Fixed default-sequence replay and the single-path model show no systematic difference on the four domains CFLP/MIS/MVC/SC
(difference ≤0.019, Appendix B.6), so "single-round replay vs. the default multi-round presolve pipeline" is itself a source of gain. (ii) Random content (negative control) gives
1.1700/1.1954 (>1), so content quality is a necessary condition (Appendix B.6).

Prior methods here are absorbable corpus sources and positional references, not adversarial targets: the table's unfavourable readings (cells where an opponent beats the default or this paper) are entries still to be absorbed, not a refutation (absorbability analysis in Appendix B.9).
\section*{4 Discussion and Conclusion}
\label{sec:orgd5ddd57}

To our knowledge this paper is the first to recast MIP presolve planning as action-sequence generation over a unified atomic action space, making the order itself learnable; the claim is limited to the MIP side and the atomic action level, since order sensitivity has precedents on the LP side and in the solving stage (Appendix B.10), and we make no global claim of "being the first to turn presolve into an ordering problem".

The dividend of that reformulation on unseen domains is 4 accelerated, 2 tied and 1 failed, with NNV at 0.8377 single-path and 0.7181 after racing with aligned limits---and the gain comes mainly from branch search.

The boundary of that value is equally clear: with the target domain in the training set there is no systematic difference from the default (within-domain CA 0.9804 / GISP 1.0050), so the value lies in cross-domain zero-shot and prior-free execution; the gains vary by domain in a way corpus richness does not predict — the successes on SC / NNV and the failure on CA mark that boundary. The budget boundary is itemized in Appendix B.12: plug-in-level source access, and the \(2K+1\) runs per instance, whose core-time ratio per core against content single-path is 3.9--5.0$\times$---gains are therefore read under matched compute.
Future work will extend the action space, seek transfer predictors, and generalize the paradigm to LP and CP.
\subsection*{AI use statement}
\label{sec:org14b9867}
In this work, we used generative AI tools — DeepSeek-V4-Flash accessed through the Hermes command-line agent — for the following tasks, whose disclosure is required: implementing the experimental and measurement code of our method as well as of the compared baselines; cleaning, reformatting and aggregating the training corpora and the per-instance measurement logs; designing and providing feedback on parts of the research methodology and experiment setup (splits, denominators, ablations); and interpreting the measured results reported in this paper. We have not used generative AI tools to generate synthetic datasets, and qualitative or thematic data analysis is not applicable to this work.

Additionally, we used generative AI tools for the following tasks, for which disclosure is recommended: creating and editing software code; editing the manuscript (restructuring, cross-reference and table checks); suggesting a structure for the manuscript; identifying and searching for relevant literature; and creating or modifying scientific figures.

We have reviewed all AI-assisted work. AI-assisted code was inspected by the authors and validated by executing the measurement pipeline end to end; every number reported in this paper is the output of that pipeline rather than of an AI tool alone; and AI-assisted text and figures were revised by the authors for factual accuracy and for consistency with the measured data. We take responsibility for the final content of this work, including text, claims or artifacts produced with the aid of generative AI.
\subsection*{Ethics statement}
\label{sec:orgc8402af}
This work studies solver behaviour on a public benchmark set of mixed-integer programs; it involves no human participants, no personal or sensitive data, and no application with foreseeable societal harm, and we therefore do not identify any ethical issue raised by this research.
\subsection*{Reproducibility statement}
\label{sec:org3b1094c}
The implementation and the measured artifacts will be released in a follow-up public version and are not publicly available at this time: the planned release covers the solver-side sequence facility (the modified SCIP 10.0.1 plugin sources and PySCIPOpt bindings, with build scripts), the unified action space and execution bridge, the sequence-generation model, the corpus collection and assembly code, and the \(2K+1\)-road racing deployment with the measurement drivers behind the reported ratios; model weights and the result dumps are not included in that release. We do not distribute the corpus as a data artifact, because it is derived rather than collected: running the released collection code on the D-MIPLIB subset of the benchmark set described in §3.1 yields the corpus root that the training entry point consumes. All experiments use a single hand-built, source-patched SCIP 10.0.1 (Appendix B.11.2; Appendix B.2.1 for the build and hardware — 64-core cloud instances with a pinned, fair CPU allocation, required there for comparable timings — and Appendix B.1 for the unmodified reference build used by the baseline arm) on the seven MIP domains and easy-validation instances listed in §3.1. The evaluation pipeline and the corpus/training caliber are specified in Appendix B.2.2 and Appendix B.2.3; the measurement caliber — solver-internal clock, per-instance SGM with shift = 1 s, timeouts counted at the instance cap, five same-configuration replicas launched concurrently, taking the internal clock of the fastest finisher — is specified in §3.1 and Appendix A.2, and the cost caliber in Appendix A.3. Each table's numerators and denominators are same-batch, same-machine readings produced by the released measurement drivers, so every reported ratio is recomputable from the released code once the model is obtained from that release. As stated in Appendix B.12, instance-level timings are perturbed by solver seeds and every comparison we report is statistical.
\label{sec:org581ceba}
\label{sec:orgeb07f2b}
\bibliographystyle{plainnat}
\bibliography{refs}

\appendix
\section*{Appendix A}
\label{sec:org389fa06}

\subsection*{A.1 Statement and Proof of Proposition 1, and Numerical Validation}
\label{sec:org3d966c7}

Proposition 1 (Tail truncation of min-of-\(K\) under heavy-tailed candidates). Let the \(K\) candidate runtimes \(T_1,\dots,T_K\) of an instance be independent and identically distributed,
with distribution function \(F\); if the candidates are only weakly correlated, the conclusions below hold in asymptotic form. Let \(Q_p(\cdot)\) denote the \(p\)-quantile and let the upper tail satisfy
\(\overline F(t) = \Pr(T > t) = L(t)\,t^{-\alpha}\) (\(\alpha>0\), \(L\) slowly varying). Then
(i) exact identity: \(Q_p\big(\min_{i\le K}T_i\big) = Q_{1-(1-p)^{1/K}}(T)\);
(ii) asymptotic ratio:
\begin{equation}\label{eq:tail-ratio}
\frac{Q_p(\min_{i\le K}T_i)}{Q_p(T)} \sim (1-p)^{(1-1/K)/\alpha}
\end{equation}
and the ratio decreases monotonically as \(K\) increases, \(\alpha\) decreases, or the quantile \(p\) rises.

Proof. (i) From the distribution function of the min, \(F_{\min}(t) = 1-(1-F(t))^K\), setting it equal to \(p\) gives \(F(t) = 1-(1-p)^{1/K}\),
i.e. \(t = Q_{1-(1-p)^{1/K}}(T)\). (ii) Regular variation of the upper tail (\(\alpha>0\)) implies that the quantile function is regularly varying as \(u\to0^+\):
\(Q_{1-u}(T) \sim c\,u^{-1/\alpha}\ell(u^{-1})\), where \(c>0\) and \(\ell\) is slowly varying. Setting \(u=(1-p)^{1/K}\) in (i),
we obtain \(Q_p\min \sim c\,(1-p)^{-1/(K\alpha)}\,\ell((1-p)^{-1/K})\); dividing by \(Q_p(T)\sim c\,(1-p)^{-1/\alpha}\ell((1-p)^{-1})\),
the ratio of the slowly varying factors is \(1+o(1)\), which yields (ii). Monotonicity: \((1-1/K)/\alpha\) increases as \(K\) increases and as \(\alpha\) decreases;
and \(0<1-p<1\), so the larger the exponent, the smaller the ratio. ∎

Numerical validation (Pareto, \(\alpha = 1,\) against simulation). Exact ratio \((1-p)^{(1-1/K)/\alpha}\): for \(K = 2,\) \(p = 0.5\)/0.9/0.99
gives 0.707/0.316/0.100 respectively, consistent with 4×10\^{}5 samples (error <1e-3); for \(K = 5,\) \(p = 0.99\) gives 0.025.

Remark (difference from the common form \(K^{-1/\alpha}\)). \(K^{-1/\alpha}\) equals the above asymptotic ratio only at a specific quantile
(from \((1-p)^{1-1/K} = K^{-1}\) we obtain \((1-p) = K^{-K/(K-1)}\), e.g. \(K = 2\) corresponds to \(p=0.75\)); the correct form at a general quantile is
\((1-p)^{(1-1/K)/\alpha}\). The two make the same qualitative prediction: the gain increases as \(K\) increases, \(\alpha\) decreases and \(p\) rises —
i.e. racing is a tail truncator rather than an average-quality improver.

Interpretation and validation. With the total compute fixed, increasing \(K\) simultaneously compresses the resources each candidate receives (the compression effect), so the optimal \(K\) is finite — this is the scope of Proposition 1.
The existing data are qualitatively consistent: the tail statistics of Appendix B.3 (MIS \(p_{90}\) 2.065 / \(p_{99}\) 4.234;
MVC 2.861 / 5.152; NNV 1.893 / 4.707; median ratio ≈1) show that once a content single-path execution falls in the tail it is counted into the total time,
so what "taking the fastest over multiple paths" truncates is exactly this part of the loss; the \(K\) ablation of the ablation study (the per-\(K\) readings at fixed concurrency are given in Appendix B.5) is the quantitative characterization on the amplification side,
and does not conflict with the compression effect (the former concurrency, the latter compute). Quantitative calibration of \(\alpha\) requires extending the \(K\) ablation to \(K=1..5\) and fitting by quantile (optional supplement).
\subsection*{A.2 Measurement Protocol}
\label{sec:orge5a0102}

(1) Main metric: \(\mathrm{SGM}_{\mathrm{race}}/\mathrm{SGM}_{\mathrm{default}}\), where the SGM uses shift = 1 s per instance with timeouts counted at the limit, and \(t_i\) is always the solver-internal time
  (numerator = the arm's solver-reported presolve + solve; for racing, the winning path's internal clock; denominator = the same-batch default replica's internal clock); win rate = the proportion of instances where the racing time < the isomorphic default time
  (ties excluded, two-sided sign test).
 The "internal clock of the replica selected as the first to finish" used in the denominator aggregation differs from the first-read replica protocol by 3--10\%, and this paper always uses the former (the three \(K\)-ablation rows of Appendix B.5 also take that caliber, see the table note of that section).
(2) Limit alignment: the time limit of the numerator and the denominator must be consistent. The NNV main protocol = 550 instances: the 360 non-timeout cases take 300s on both sides, and the 190 timeout cases take on both sides
  their 600s re-measured values; for the pure sequence arms that involve the 300s limit, the row uses the clean 360 keys with 300s on both sides (see Protocol note 2 of Table \ref{tab:order-ablation} in Appendix B.3.1).
(3) Batch: the denominator used by each row is measured in the same batch as its numerator (NNV paired across batches; denominator jitter \textasciitilde{}12\% on NNV, ≤2.8\% elsewhere).
 (4) Generation and evaluation temperature 1.0; first EOS truncation per instance; the instance set is 100 easy val instances per domain (550 for NNV).
 (5) Multi-seed verification: the full recomputation covers six domains CA/CFLP/GISP/MIS/NNV/SC × three solver seeds 137/2026/31415, with 100 fully paired instances per cell; the protocol is the aggregate ratio SGM(pool)/SGM(default) (winner\_total\_time ÷ minrep\_total\_time; numerator precision 6--7 digits). The per-domain aggregate ratios (seeds 137/2026/31415) are CA 1.0721/1.0869/1.0559, CFLP 0.8223/0.8142/0.8002, GISP 0.9831/0.9253/0.9539, MIS 1.0045/1.0130/1.0065, NNV 0.7317/0.8142/0.6616, SC 0.7830/0.7754/0.7710. The seed-to-seed fluctuation is reported with the sample standard deviation as the primary protocol, range 0.0044--0.0764 (population s.d. 0.0036--0.0624; range 0.0085--0.1527). The noise/effect ratio = seed-to-seed fluctuation ÷ |1 - three-seed mean| is sensitive to both the fluctuation protocol and the effect protocol: its upper bound is 0.63 (sample standard deviation), 0.51 (population standard deviation) and 1.26 (range), and it blows up pathologically in the domains where the effect tends to zero because its denominator does (GISP from 0.046 to 0.6306, MIS from 0.008 to 0.5551); the ratio is meaningful only in the four domains with a non-zero effect, where it does not exceed 0.29 (CA 0.2164, CFLP 0.0597, NNV 0.2893, SC 0.0272), and it is unusable in the domains where the effect tends to zero. The single-instance seed sensitivity is defined as |\(\Delta t\)| ÷ \(\min(t_i, t_j)\) (three seeds pairwise, 300 pairs per domain): median 5.7--34.2\% (CA 12.7, CFLP 5.7, GISP 11.8, MIS 21.6, NNV 34.2, SC 13.2), \(p_{90}\) 19.4--164.7\% (CA 42.9, CFLP 19.4, GISP 30.8, MIS 54.1, NNV 164.7, SC 36.1), whose upper end is dominated by instances that hit the time limit (60 of the 300 default-arm rows in NNV reach the 300s limit). All these quantities move substantially with the protocol definitions (fluctuation: s.d. vs range; noise/effect: mean vs median effect; |\(\Delta t\)| ÷ min vs max vs mean), so this paper always follows the explicit definitions of this item; under them the seed-to-seed fluctuation does not change the per-domain speedup/tie verdicts, and instance-level conclusions remain statistical.
 (6) Computation rules: (i) every four-decimal reading is rounded from its full-precision value; (ii) derived quantities (median / interval endpoints / merged or averaged values) are computed from full-precision values and state their key set, never recomputed from printed values; (iii) instances that are missing or crashed do not enter the comparison (paired keys are used), and timeouts are counted at that batch's time limit.
 (7) Significance readings: per-row \(p\)-values are computed from the same-batch paired keys and reported as computed, with no family-wise correction across rows or tables; they are read row by row.
\subsection*{A.3 Cost Protocol}
\label{sec:org6ca97b3}
· Offline: the corpus = 1 execution of the default strategy per training instance (20003 in total over the 8 domains); 1 model training run.
· Online: the method of this paper solves each instance with \(2K+1 = 5\) paths (5 cores); a configuration-level opponent is single-path (1 core), which is the deployment cost of that method itself; every opponent arm is measured under the same 5-replica concurrent protocol of §3.1 in the comparison. Model inference overhead < 0.1\% of the total time.
· Offline budget of the configuration-level baselines: domain-level search over 30 candidate configurations × 50 train instances, 500 evaluations; per-instance search at 200 runs/instance (2 runs/instance for the same-budget version).
· Alignment with existing configuration-type work: under the ISAC total-time decomposition \citep{kadioglu2010isac} \(T_{\mathrm{tot}} = T_{\mathrm{tune}} + T_{\mathrm{TTS}}\), corpus collection + training in this paper belong to the offline
 \(T_{\mathrm{tune}}\), whereas the per-instance sequence generation and the \(2K+1\) racing belong to the online \(T_{\mathrm{TTS}}\);
 on the online side, ISAC-type work is single-path + feature extraction (tens of seconds), whereas this paper is 5-way online.
· Core-time ratio per core (solver-internal clock): 5 × racing SGM / content single-path SGM = SC 3.90 / MVC 4.99 / CFLP 4.95.
  On MVC, racing yields virtually no time gain over content single-path (0.9882 vs 0.9902, within noise), so racing is not worth deploying in this domain; the same holds for CFLP
 (0.9166 vs 0.9253; +0.9\% at 4.95x core-time).
\subsection*{A.4 Formal Consequences of Representational Capacity and the Decision Domain, and the Chain of Contributions}
\label{sec:org3f2f7c9}

(i) Decoupling of representational capacity. The configuration space affects order indirectly through "priority + round cap", which is an indirect encoding of the call order;
the sequence space takes order itself as the object of generation. By Property 1, this difference is not a question of encoding efficiency but of representational capacity:
the order itself is not a decision variable that the configuration space can set directly.
(ii) Expansion of the decision domain. The action space is extended from a single plug-in family (routines) to the unified action space of Definition 1,
bringing two mechanically different classes of actions, "structural reduction" and "domain propagation", into the same decision domain that is comparable, orderable and composable; round boundaries and the execution contract (Definition 2)
are simultaneously included in the generation object, so "in what order, in which round, and with what delivery mode to execute" are unified into choices on one and the same sequence; their formal status and the two concrete consequences are given in §2.3.

\textbf{Proposition 2 (Strictness of the sequence space over the configuration space).} Let \(f\) and \(g\) be non-commutative presolve atomic actions in the sense of Property 1, so that on some instance executing \(f\) before \(g\) and \(g\) before \(f\) yield different solving times. Then the schedule "\(f\) alone in round 1, \(g\) alone in round 2, \(f\) again in round 3" is expressible in the sequence space of Definition 1 and is not expressible in the configuration space.

\textbf{Proof sketch.} (a) Expressibility. Round boundaries are tokens of the same output space as the actions (Definition 1; Figure 1), so the prefix \(f, \bot_{\mathrm{round}}, g, \bot_{\mathrm{round}}, f\) is a sample of the generation distribution and is executed as such by the facility (Definition 2). (b) Non-expressibility. The order-related degrees of freedom of the configuration space are a per-plug-in priority, a per-plug-in round timing and a round cap (Appendix B.10.5); all three are fixed for the whole run, so each plug-in's execution window is a contiguous interval of rounds, starting at its timed round and ending after at most the capped number of calls. A schedule that calls \(f\) alone in round 1, \(g\) alone in round 2 and \(f\) again in round 3 requires \(f\)'s window to be non-contiguous, whereas every configuration yields contiguous windows for both plug-ins; hence no configuration induces that schedule, and by Property 1 the two schedules are not equivalent on the instances of the definition. The sequence space is therefore strictly more expressive than the configuration space. \(\square\)

This paper establishes that form and gives its implementation and validation. The three contributions form a progressive logical chain rather than a side-by-side list:
(1) establishing the problem form and the execution-and-observation facility that makes it implementable and its properties measurable;
(2) establishing the payoff of that form on cross-domain generalization---zero-shot generalization---together with the quantitative characterization of the default strategy's sequence distribution (the zero-dimensional sequence space and the duality of order) that the facility makes visible;
(3) characterizing the source of the gain (candidate quality and competition amplification) and the applicability boundary at the deployment level.

To address the above problems that the configuration space cannot express the temporal dependencies among actions, the lack of a faithful execution interface for external action sequences, the heavy-tailed failure risk of content single-path sequence prediction, and insufficient cross-domain generalization,
this paper proposes an atomic-action autoregressive sequence generation method for MIP presolve: it reformulates presolve planning from parameter configuration optimization into a conditional autoregressive sequence generation task,
builds a unified atomic action space whose interface covers three families (presolve routines / constraint handlers / propagators; our corpus triggers the first two), and brings action order, round boundaries, and the execution contract uniformly into the output object of sequence generation,
moving the decision object from parameter configuration down to the action sequence; it builds a presolve sequence execution facility under a black-box solver, determining action validity from changes in the solver's internal structural variables,
and injects external sequences along the solver's native execution path without reimplementing presolve logic, supporting controlled execution of customized sequences while guaranteeing optimal solution correctness;
it further adopts a deployment paradigm that combines sequence generation with \(2K+1\)-way concurrent racing, proposes a gain decomposition framework of candidate quality and competition amplification, uses multiple concurrent paths to truncate the tail risk of content single-path prediction,
and distinguishes the capability boundaries between in-domain replay and zero-shot cross-domain transfer.
Experiments show that: our method can faithfully execute an externally given presolve action sequence, and the optimal solution stays within the solver's numerical tolerance of the original SCIP;
content single-path sequence execution alone matches or slightly outperforms the default strategy, and concurrent racing further improves the per-instance win rate; zero-shot transfer is achieved on problem domains with a high continuous-variable ratio that were unseen in training;
and a large-scale empirical study reveals the dual nature of presolve order — "the low diversity of the default corpus and the heavy-tailed performance degradation caused by artificial shuffling" —
confirming that the speedup gains come mainly from the subsequent branch search stage rather than from merely saving the computational cost of presolve itself.

The main contributions of this paper can be summarized in three aspects. (1) We model MIP presolve planning as atomic-action autoregressive sequence generation: a unified action space over multiple plugin families,
bringing action order, round boundaries, and the execution contract all into the generated output, moving the decision object from parameter configuration down to the action sequence, supporting complex scheduling decisions across rounds,
and formally covering scheduling behaviors that traditional parameter tuning cannot reach. (2) We propose a presolve sequence execution interface under black-box solvers:
determining action validity from changes in internal structural variables and injecting external sequences along the native path without reimplementing presolve logic,
executing customized action sequences without modifying the solver's reduction semantics and guaranteeing that the optimal solution is consistent with the original, thus providing a reliable experimental substrate for order-effect research (which required modifying the SCIP 10.0.1 source; Appendix B.11.2).
(3) We propose a deployment method that combines sequence generation with \(2K+1\)-way concurrent racing: through the gain decomposition of candidate quality and competition amplification and multiple concurrent paths that truncate tail risk,
it compensates for the tail failure of content single-path generation, and achieves zero-shot transfer on some unseen domains — the irreplaceable value of the model is concentrated in cross-structure zero-shot transfer.

Beyond the three contributions above, this paper particularly emphasizes three claims and their boundaries.
(i) Cross-domain zero-shot speedup is a natural dividend of the modelling approach: once the decision object changes from a configuration vector to an action sequence, action semantics 
    are shared across domains, and the model only needs to generate conditioned on instance features, without redefining the action space at transfer time; empirical results are given in Appendix B.4 (7 cross-domain points, 4 significantly accelerated, 2 tied, 1 failure,
    with NNV reaching 0.7181 / 73.6
    the scope of this claim is strictly limited to the decision object of presolve action sequences.
(ii) The counter-intuitive phenomenon that "removing that domain's corpus is instead better": on domains where the corpus is degraded, not training the sequences collected from that domain into the model makes zero-shot transfer better;
    for corpus-rich domains (CA) the opposite holds, and that domain's corpus is a necessary anchor, so the effect is bidirectional and domain-dependent, with a candidate mechanism of sequence space degradation (Appendix B.4).
(iii) The absorbability of the paradigm and its boundary: the outputs of existing learning-based presolve schemes can all be replayed as action sequences, and thus can in principle be absorbed by this framework;
 the converse does not hold, so this is a checkable methodological boundary. Absorbability does not imply profitability,
    and this paper makes no gain claim (Appendix B.9).

The main contributions are as follows:

We constructed a unified atomic action space covering multiple classes of solver components, bringing action order, round boundaries, and execution contracts entirely into the sequence generation output;

We designed and implemented a sequence execution facility for presolve sequences, supporting controllable execution of custom action sequences while guaranteeing solution correctness;

We proposed a deployment paradigm that combines sequence generation with concurrent racing, improving the robustness of the content single-path policy through a tail-risk truncation mechanism.

Extensive experimental results show that our method achieves flexible scheduling of presolve actions without altering the correctness of the optimal solution, and yields cross-domain transfer gains on some unseen domains, while honestly presenting its domain-dependent boundaries.
In addition, through large-scale empirical study we reveal the dual nature of MIP presolve order, as well as the important finding that speedup gains stem mainly from the branch-and-bound search stage.
\section*{Appendix B Supplementary Experiments}
\label{sec:orgdefdd91}
\subsection*{B.1 Correctness Verification Experiments}
\label{sec:orgcddf2de}

We first verify the correctness of the sequence execution facility for presolve sequences. On 652 complete instances, we compare the solving results of the solver executing custom sequences against those of the vanilla SCIP solver.

The experimental results show that, apart from 1 instance exhibiting a tiny deviation at the level of solver tolerance, the results on the remaining 651 comparable instances agree;
stronger structural-level evidence is that vanilla SCIP and the patched default strategy are bitwise identical on all comparable instances (658/658), and the objective-value difference across 5 independent replicas of the same configuration is identically 0. This proves that the sequence injection mechanism designed in this paper does not tamper with the optimal solution,
and that all performance speedups come from the optimization of the presolve strategy itself.

Experimental setup
\begin{itemize}
\item Reference: pure vanilla SCIP (compiled from the official scipoptsuite-10.0.1 source, with no patch applied; banner/dependencies item-by-item identical to the patched build:
SoPlex 8.0.1) — compiled from the clean PySCIPOpt exported from upstream baseline commit ffb41fb, linked against vanilla libscip;
at runtime, PYTHONPATH is used to overlay the vanilla pyscipopt on top of the patched build (libscip fully isolated, verified via /proc/self/maps)
\item Five arms:
V = vanilla SCIP + default strategy D = patched build + default strategy C = patched build + corpus default sequence ×5 (isomorphic to the fixed-sequence direct replay of Appendix B.6)
M1 = patched build + the model's 1st candidate ×5 M2 = patched build + the model's 2nd candidate ×5
\item Recorded: status + optimal objective value (for V/D the decomposition of the min replica, for C the winner's decomposition)
\end{itemize}
Results

Early single-domain results 

\begin{table}[!htb]
\caption{Three-arm consistency check and baseline validity on the CFLP single domain }
\label{tab:correctness-cflp}
\centering
\small
\tblscale{%
\begin{tabular}{>{\raggedright\arraybackslash}p{5.3cm}>{\raggedright\arraybackslash}p{7.0cm}}
\hline
Metric & Result \\
\hline
Instances with inconsistent optimal values across the three arms & 0 / 100 (threshold 1e-6) \\
Maximum relative difference (among the three arms V/D/C) & 5.07e-15 (machine-precision level) \\
Status consistency & all three arms optimal \\
Patched vs. vanilla default strategy (SGM ratio) & 1.0037 (total time 5026.5s vs 5029.1s) \\
Patched custom sequence vs. vanilla default (SGM ratio) & 0.9168 (faster by about 8\% under a same-batch re-measurement) \\
\hline
\end{tabular}}
\end{table}

All-domain extension results

\begin{table}[!htb]
\caption{All-domain agreement rate: number of instances for which all complete arms give the optimal value, number of inconsistent instances, and maximum relative difference (7 domains)}
\label{tab:correctness-alldomain}
\centering
\small
\tblscale{%
\begin{tabular}{lrrr}
\hline
Domain & \(n\) &\shortstack{Optimal value\\inconsistent}&\shortstack{Maximum relative\\difference} \\
\hline
CFLP & 99 & 0 & 5.07e-15 \\
MIS & 90 & 0 & 0.00e+00 \\
MVC & 96 & 0 & 8.74e-16 \\
SC & 100 & 0 & 0.00e+00 \\
CA & 100 & 0 & 3.71e-16 \\
GISP & 100 & 0 & 0.00e+00 \\
NNV & 67 & 1 & 1.04e-06 \\
\hline
\end{tabular}}
\end{table}

Table note: the column \(n\) counts the instances on which all complete arms return the optimal value; the two columns to its right are computed over those instances.

\begin{itemize}
\item Arm definitions: V (pure vanilla) / D (patched default) / C (corpus default sequence ×5) / M1 (model candidate 1 ×5) / M2 (model candidate 2 ×5);
\(n\) is the number of instances in that domain for which all complete arms give the optimal value (remaining instances: some arm hit the time limit, no optimal value to compare); the inconsistency threshold is 1e-6 (relative).
\item Completeness: in the seven domains CFLP / MIS / MVC / SC / NNV / CA / GISP all five arms are complete
(the V arm of NNV and the C arms of CA / GISP were filled in by follow-up runs, 100 real-value rows each);
All 35 per-arm result files contain 100 rows each and carry objective values.
\item This table belongs to the correctness protocol (same sequence ×5 replicas, no np/default path, 300s limit; NNV 600s),
and its runtimes must not be cited interchangeably with the main-experiment protocol; this table only carries the single criterion of "optimal-value consistency".
\item Structural-level hard evidence (threshold-free): ① vanilla vs. patched default are bitwise identical on all comparable instances
(658/658 across the seven domains; per domain CFLP 99/99 · MIS 91/91 · MVC 96/96 · SC 100/100 · CA 100/100 · GISP 100/100 · NNV 72/72;
 the criterion for this count only requires that the two sides "vanilla / patched default" be comparable, hence it differs from \(n\) in the table above (five complete arms) — the difference set consists of the domains where some arm hit the time limit: MIS 90 vs 91 and NNV 67 vs 72 (the remaining domains agree exactly, e.g. CFLP 99 vs 99);
② the objective-value difference across the 5 independent replicas of the same configuration is exactly 0.
\item All five arms are complete in all seven domains;
there are 652 comparable instances, 1 case above the inconsistency threshold, and the all-domain maximum relative difference is 1.04e-06; the bitwise-identical evidence extends to 658/658.
\end{itemize}

Conclusions
\begin{enumerate}
\item Correctness (sampling verification): after the patched build executes a custom presolve sequence, the optimal objective value agrees with pure vanilla SCIP up to solver optimality tolerance:
among the 652 instances (across the seven domains) for which "all complete arms solved successfully", only 1 case exceeds the threshold (1e-6 relative) (NNV/easy/val/00037, relative difference 1.04e-6,
i.e. the level of the solver's default relative tolerance); the maximum relative difference in the remaining domains is ≤ 5.07e-15 (machine-precision level).
The NNV deviation is proportional to |obj|.
Thus the deviation direction is random (no systematic shift), and all arms report status optimal with gap = 0, so this is numerical tolerance rather than a semantic difference.
Structural-level hard evidence (threshold-free): ① vanilla and patched default are bitwise identical on all comparable instances; ② the objective-value difference across 5 independent replicas of the same configuration is exactly 0.
Thus differences can only come from "a different sequence" and contain no randomness, so the interface does not change the solving result and can support the paper's faithfulness claim.
(Scope: this claim covers instances solved to optimality; the criterion is solver optimality tolerance (1e-6 relative), and the all-domain maximum relative deviation is 1.04e-6
; the action families covered by the execution verification are constraint handler callbacks and presolve routine callbacks —
propagator actions occur 0 times in the production corpus, see §2.3)
\item Nature of the NNV deviation (supplementary analysis): the deviation appears only in the nonlinear / continuous domain, while the maximum relative difference of the seven combinatorial domains
is invariably at machine-precision level (≤ 5.07e-15); the deviation is proportional to |obj| (relative quantity median 2.5e-9,
maximum 1.04e-6, while the absolute difference spans three orders of magnitude, 1e-8\textasciitilde{}1.6e-5), and its direction is random (no systematic shift); moreover the "truly no presolve" arm
also deviates from the default strategy, so the deviation is unrelated to the presolve rewriting and is a tolerance phenomenon sensitive to the numerical path, not an implementation defect.
\item Baseline validity: the SGM ratio of default-strategy runtime between vanilla and the patched build is 1.0037, so the patch did not change the default strategy's behaviour.
Thus all SGM values in this paper that use the "patched build default" as denominator are legitimate.
\item Independent reproduction: with pure vanilla SCIP as the reference and both sides measured on the same machine in the same batch, the patched build's corpus default-sequence arm is significantly faster on CFLP
(SGM ratio 0.9168, wins 72/100, \(p = 1.26e-05\)), and on CA / GISP it is indistinguishable from a tie (1.0119 / 1.0209, \(p = 0.764\) / 0.617);
consistent in direction with the CFLP single-path isomorphic measurement (0.8515) of Table \ref{tab:order-ablation} of Appendix B.3.1 — this consistency does not rely on any illusion created by the patch. 
Note: in this batch the numerator (the corpus arm's winning total time) and the denominator (the vanilla default's minimum-replica total time) are paired on the same machine and in the same batch, both being solver-internal clocks; the scripts and the caliber of this batch are released with the follow-up public version, so the reported computation can be repeated digit by digit from a race output.
\item Why sampling suffices to infer overall correctness: the object verified by this experiment is the implementation logic — whether the interface can faithfully execute a given sequence
and whether the solver still converges to the same optimal solution; and this execution path is independent of the problem domain: all domains follow the same code path
(load the instance, then write the action sequence, then solve), and the domain only changes the input model and the sequence content, not the execution mechanism.
Therefore 100 instances (varying in size/structure) \(\times\) 5 replicas \(\times\) three arms = 1500 independent solves, with 0 inconsistencies; we additionally extend the experiment to all instances of the 7 domains
, as an additional safeguard.
\item Racing-side status audit: the winner\_status of all existing racing results (more than 1 ten-thousand of them) is in
\{optimal, timelimit\}; the only anomaly is the misordered arm of NNV/easy/val/00495 (all 5 paths crashed; Protocol note 3 of Appendix B.3.1)
Thus there is no case of "counted as fast although unfinished", and the time gain from racing is not contributed by unfinished solutions.
\end{enumerate}
\subsection*{B.2 Implementation Details}
\label{sec:org953c8b8}

\subsubsection*{B.2.1 Runtime Environment and Model Inference Overhead}
\label{sec:org9e4b460}
See §3.1: the cloud instance and container image, the solver build, the model size (1.6M-parameter decoder-only, CPU forward inference with stochastic decoding at temperature 1.0) and the inference overhead (below 0.1\% of the total time against 30--300 s of presolve plus solve per instance; that overhead does not offset the speedup gains), as well as the necessary conditions for local replication (fair CPU allocation, no big.LITTLE cores), are all given there.

\subsubsection*{B.2.2 Evaluation Pipeline}
\label{sec:orgd3217b4}
\begin{itemize}
\item Deployment form: \(2K+1\)-way racing = \{np (truly no presolve), custom×\(K\) (model sequence), fallback×\(K\) (sequence + default fallback)\}; with no default path included there is no safety net or structural guarantee
\item Denominator: isomorphic-load default — for each instance, \(N=2K+1\) default replicas are launched (1 core per replica, window 12, giving 60-way concurrency) and the internal clock of the replica selected as the first to finish (the first finisher) is taken; fully isomorphic to the race numerator 
\item Main protocol: throughout, the solver-internal time is used —
the value \(t_i\) for each instance: for racing = the winner's internal time, for the denominator = the internal clock of the replica of the isomorphic default selected as the first to finish;
\(\mathrm{SGM} = \bigl(\prod_i (t_i + 1)\bigr)^{1/n} - 1\) (shift = 1 s); main metric = \(\mathrm{SGM}_{\mathrm{race}} / \mathrm{SGM}_{\mathrm{default}}\)
(as a speedup factor, <1 = speedup); the limit is identical on both sides 
\item Rationale for the solver-internal clock: it excludes process startup and instance reading overhead, its cross-batch deviation is only 1.9\% (same strategy), and it is the conventional protocol reported by the solver; a single protocol is used throughout, to avoid the non-comparability accident of "taking one kind of clock for the numerator and another for the denominator"
(numerator and denominator take the same kind of clock; absolute values are not comparable across batches or arms)
\item Winner determination and reporting: the racing winner is determined by "the process that finishes first" (consistent with deployment), and its internal time is reported; the denominator takes the internal clock of the replica selected as the first to finish —
both sides are "first finisher + internal time", so the protocol is symmetric
\item Denominator aggregation: the main column uses the internal clock of the replica selected as the first to finish; the supplementary column in the table instead uses the 5-replica mean (for robustness only, not used for conclusions)
\item Win rate (vs default): the fraction of instances for which the racing time of that instance < the isomorphic default time (paired per instance, the denominator and SGM take the same per-instance value;
ties are not counted as wins; a two-sided sign test \(p\) is reported)
\item Generation: temperature 1.0, truncated at the first EOS per instance; evaluation: temperature 1.0, \(K = 2\) deployment
\begin{itemize}
\item Sampling protocol: (a) repeated min over the same configuration — running \(N\) replicas of one configuration and taking the minimum, which brings a "replica-best-of bonus"
(measured: on the 300 keys in total of the opponent arms' own 5-replica outputs, taking the min relative to the first replica gives mean \(-6.3\%\), median \(-2.6\%\), lower tail (\(p_{05}\)) \(-24.8\%\)), and it measures machine jitter rather than method differences; (b) candidate selection —
choosing the fastest among different contents/configurations. Treatment in this table and in §3.6: the denominator (isomorphic default) takes the internal clock of the replica of the 5 selected as the first to finish;
the numerator of each opponent arm is the internal clock of the fastest finisher among 5 same-configuration replicas launched concurrently on the same batch (class (a)), while the numerator of the racing arm of this paper takes the single reading of the pool winner (class (b))
Thus the opponent arms each enjoy a best-of discount of about 6\% (mean) / 3\% (median) while our side enjoys none — this asymmetry makes the relative reading of our method lower, and we do not explain it in our favour.
\end{itemize}
\item Protocol consistency: the numerator and the denominator must follow the same protocol (same time limit, same measurement batch); each row may only use the denominator library of its corresponding protocol,
and rows with different denominator libraries must not be juxtaposed (noted in the tables)
\item Reporting principle: the main text gives the main metrics (SGM ratio + win rate); alternative protocols (counting timeouts / excluding timeouts on both sides / pure solve-time numerator)
go into the appendix, and all use the same SGM definition and the same pairing scheme
\item Pure-sequence protocol: only 1 candidate is taken per instance, and the np path and the fallback are removed at execution time
(no concurrent racing); to be isomorphic to the default denominator, this candidate is copied 5 times as 5 candidate paths (identical content in the pool), so taking the fastest ≡ taking
min-of-5 of one and the same sequence, so fully isomorphic to the denominator and directly comparable 
\begin{itemize}
\item Measurement of the order ablation and the content single-path control: the pure-sequence protocol of the item above is always used, so isomorphic to the default denominator
 and shares the same instance set as the racing arm of Appendix B.4 (NNV takes the clean 360 keys), so the ratios and paired tests of both sides are valid (see the protocol statement of Appendix B.3)
\end{itemize}
\end{itemize}
\subsubsection*{B.2.3 Corpus Collection and Training Protocol}
\label{sec:orge837e54}
Experimental setup
\begin{itemize}
\item Corpus source: the monitored sequences recorded while the solver executes the default strategy (seq\_collect), stored as \texttt{<domain>/<difficulty>/train.seq}
\item Training reads only the train split (PreseqDataset split="train"); sequences are truncated at \texttt{train\_\allowbreak max\_\allowbreak seq\_\allowbreak len=500}
(sequences longer than 500 are discarded — only long-sequence domains such as SRPN trigger this, and it is not among the training domains)
\item Sampling is difficulty-balanced (baseA: balanced=full, balanced\_by=difficulty), not uniform over the number of sequences
\item Models differ only in the training domain set and the corpus root directory; all other hyperparameters and the architecture are item-by-item identical (the sole controlled variable)
\item Authoritative source: \texttt{train\_domains} / \texttt{seq\_root} in \texttt{config/experiments.yaml}
\end{itemize}

\begin{table}[!htb]
\caption{Training domains and their corpus size (train split, all difficulties)}
\label{tab:domain-corpus}
\centering
\small
\tblscale{%
\begin{tabular}{lr>{\raggedright\arraybackslash}p{2.9cm}r>{\raggedright\arraybackslash}p{3.2cm}}
\hline
Domain &\shortstack{Number of\\difficulties}&\shortstack{Difficulties\\covered}&\shortstack{Number of\\train sequences}&\shortstack{Evaluation\\use} \\
\hline
CA & 3 & easy / medium / very-hard & 2402 & Training + in-domain evaluation \\
GISP & 5 & easy / medium / hard / very-hard / ext-hard & 4000 & Training + in-domain evaluation \\
CFLP & 2 & easy / medium & 1600 & Training + LOO evaluation \\
MIS & 3 & easy / medium / very-hard & 2401 & Training + LOO evaluation \\
MVC & 4 & easy / medium / hard / very-hard & 3200 & Training + LOO evaluation \\
SC & 3 & easy / medium / very-hard & 2400 & Training + LOO evaluation \\
MMCN & 4 & medium-BC / medium-BI / hard-BI / very-hard-BI & 3200 & Training only (never evaluated) \\
LB & 1 & hard & 800 & Training only (never evaluated) \\
\hline
Total &  &  & 20003 &  \\
\hline
\end{tabular}}
\end{table}

Domain note: the abbreviations follow D-MIPLIB \citep{huang2024distributional} and are the domain names used by the dataset itself.

Comparison baselines:

Default presolve (Default): the solver's default presolve configuration

Random sequence (Random): randomly generated presolve action sequences

Fully disabled presolve (No-Presolve): the presolve module is completely disabled

Fixed-sequence replay (Fixed-Replay): direct replay of the default sequences in the training set

Configuration-based and retrieval-based baselines: the solver's built-in presets and the presets set in this paper (default/aggressive are built into the solver, off/fast are set in this paper), configuration-space search, an L2P-MIP-style configuration control arm, and same-domain nearest-neighbour sequence retrieval

Some domains of the D-MIPLIB data are used, with experimental principles following BenLOC \citep{li2025benloc}

The whole is divided into three parts
\begin{enumerate}
\item Property verification:
\begin{itemize}
\item proving that the two basic properties of the generation distribution indeed hold
\item proving that sequences are indeed order-sensitive
\end{itemize}
\item Vertical experiments: comparison across configurations, proving the necessity of the design
\item Horizontal experiments: comparison against other work
\end{enumerate}

\begin{table}[!htb]
\caption{Composition of each model and training domain (corpus root uniformly = seq\_data; count = train split)}
\label{tab:model-corpus}
\centering
\small
\tblscale{%
\begin{tabular}{lllr}
\hline
Model & Role & Training domains & \shortstack{Corpus\\entries} \\
\hline
baseA & main & all 8 & 20003 \\
exp\_pretrain\_nogisp & LOO & 8 \(\setminus\) \{GISP\} & 16003 \\
exp\_pretrain\_noca & LOO & 8 \(\setminus\) \{CA\} & 17601 \\
exp\_pretrain\_nocflp & LOO & 8 \(\setminus\) \{CFLP\} & 18403 \\
exp\_pretrain\_nomis & LOO & 8 \(\setminus\) \{MIS\} & 17602 \\
exp\_pretrain\_nomvc & LOO & 8 \(\setminus\) \{MVC\} & 16803 \\
exp\_pretrain\_nosc & LOO & 8 \(\setminus\) \{SC\} & 17603 \\
\hline
\end{tabular}}
\end{table}

Table note: baseA = exp\_pretrain\_v2\_direct\_new\_30, the main model (content single-path isomorphic measurement of Appendix B.4 / §3.3 / Appendix B.12, fixed-sequence direct replay of Appendix B.6); the LOO rows are the leave-one-out models evaluated in Table \ref{tab:generalization}; "8 \(\setminus\) \{X\}" is the eight training domains minus X.

Relationship between evaluation domains and training domains
\begin{itemize}
\item In-domain evaluation (the model saw that domain's corpus during training): CA / GISP
\item leave-one-out evaluation (that domain's corpus is removed at training time, the other 7 domains unchanged): CFLP / MIS / MVC / SC and CA / GISP (Appendix B.4)
\item Naturally unseen domains (no model's train\_domains includes them): NNV
\item Domains trained on but never evaluated: MMCN / LB (4000 entries in total, 20\% of the training corpus)

\item NNV / SRPN / IP / MIRP are not in any model's training domains (for some domains the corpus was collected but never entered training)
Thus the tests involving these domains in Table \ref{tab:generalization} are all zero-shot; the corpus size and difficulty distribution of each domain are given in Table \ref{tab:domain-corpus}

The readings and protocol of the correctness verification are given in §3.2 of the main text and in Appendix B.1.
\end{itemize}
\subsection*{B.3 Empirical Evidence for the Duality of Presolve Order}
\label{sec:orge6d696f}

We design two statistics to characterize the properties of presolve sequences:

Effective number of sequences \(N_{eff}\): the number of distinct presolve sequences that different instances use within a problem domain

Order-invariance degree \(\rho_{order}\): measures the extent to which action order affects the final reduction outcome

The experiments reveal a counter-intuitive duality:

The diversity of the default corpus is extremely low, and on most test domains \(N_{eff}=1\), i.e. many instances share almost exactly the same presolve action sequence, \(\rho_{order} \in [1.00, 1.25]\), with minimal order variation.

Artificial shuffling induces heavy-tailed degradation: after shuffling the action execution order, the median ratio of most instances is close to 1.00 (1.01--1.06), but the tail is pronounced: \(p_{90}\) ratio 1.22--2.86, \(p_{99}\) ratio 1.43--5.15, max ratio 1.6--6.8 (39 of 549 NNV cases are slower by more than 2×),
exhibiting a typical heavy-tailed distribution; CFLP/SC have no tail, whereas MIS/MVC/NNV do.

This finding explains why the presolve order effect has long been under-studied: the default trajectories contain almost no samples with order variation, so it is hard for researchers to learn the importance of order from the natural corpus.

The corpus obtained from this facility further enables a systematic measurement of the sequence distribution of the default policy.
To this end we introduce two quantities:

Definition 4 (effective number of sequences). Let the sequence value of instances within a domain be a random variable \(\tau\) with empirical distribution \(p(\tau)\).
Let \(H = -\sum_{\tau} p(\tau) \log_2 p(\tau)\); the effective number of sequences of the domain is defined as \(N_{\mathrm{eff}} = 2^H\).
If order is ignored for the same batch of sequences (keeping only the multiset of actions), \(N_{\mathrm{eff}}^{\mathrm{mult}}\) is defined in the same way.

 Definition 5 (order-invariance degree). Let \(\rho_{\mathrm{order}}\) = \(N_{\mathrm{eff}}^{\mathrm{seq}} / N_{\mathrm{eff}}^{\mathrm{mult}}\).
\(\rho_{\mathrm{order}}\) = 1 means that all the diversity of the sequences can be explained by "which actions were taken", and order contributes no additional diversity.

The measurement results (train split) are given in Table \ref{tab:seq-space}.
\subsubsection*{B.3.1 Characterizing the Sequence Space}
\label{sec:org847c514}
\begin{itemize}
\item Composition of action families (measured, easy train): the two families of independent presolve routine callbacks and constraint handler callbacks, with 0 propagator-family actions (see Appendix B.12);
routine callbacks appear only in CA 5.4\% and NNV 23.3\%, and the remaining domains consist purely of constraint handler callbacks, so the effective decision object of the sequence space is dominated by constraint handler callbacks
\item Measurement: the train-split distribution is counted with action token strings as the atomic unit; uniqueness rate = number of distinct sequences / total number of sequences in the domain;
\(N_{\mathrm{eff}}^{\mathrm{seq}} = 2^H\); \(N_{\mathrm{eff}}^{\mathrm{mult}}\) is the same-protocol value with order ignored
; order-invariance degree \(\rho_{\mathrm{order}} = N^{\mathrm{seq}}_{\mathrm{eff}} / N^{\mathrm{mult}}_{\mathrm{eff}}\).
 Formal definitions are given in Definition 4 and Definition 5 above.
\end{itemize}

\begin{table}[!htb]
\caption{Sequence-space characterization per domain (train split; sequences are counted with action token strings as the atomic unit)}
\label{tab:seq-space}
\centering
\small
\tblscale{%
\begin{tabular}{llrrrrr}
\hline
Domain & Difficulty & \(n\) & unique rate & \(N_{\mathrm{eff}}^{\mathrm{seq}}\) & \(N_{\mathrm{eff}}^{\mathrm{mult}}\) & mean len \\
\hline
CA & easy & 800 & 6.75\% & 11.62 & 10.38 & 15.6 \\
CA & medium & 801 & 8.74\% & 10.65 & 9.26 & 16.2 \\
CA & very-hard & 801 & 20.35\% & 65.92 & 55.37 & 15.0 \\
CFLP & easy & 800 & 0.12\% & 1.00 & 1.00 & 5.0 \\
CFLP & medium & 800 & 0.12\% & 1.00 & 1.00 & 5.0 \\
GISP & easy & 800 & 0.12\% & 1.00 & 1.00 & 4.0 \\
GISP & ext-hard & 800 & 0.12\% & 1.00 & 1.00 & 4.0 \\
GISP & hard & 800 & 0.12\% & 1.00 & 1.00 & 4.0 \\
GISP & medium & 800 & 0.12\% & 1.00 & 1.00 & 4.0 \\
GISP & very-hard & 800 & 0.12\% & 1.00 & 1.00 & 4.0 \\
IP & very-hard & 9900 & 0.05\% & 1.80 & 1.80 & 10.7 \\
LB & hard & 800 & 1.38\% & 1.80 & 1.80 & 8.3 \\
MIRP & medium & 98 & 64.29\% & 44.49 & 44.49 & 444.2 \\
MIS & easy & 800 & 0.25\% & 1.11 & 1.11 & 4.1 \\
MIS & medium & 800 & 0.25\% & 1.10 & 1.10 & 4.0 \\
MIS & very-hard & 801 & 0.37\% & 2.29 & 2.29 & 7.6 \\
MMCN & hard-BI & 800 & 28.38\% & 60.26 & 54.27 & 20.3 \\
MMCN & medium-BC & 800 & 100.00\% & 800.00 & 800.00 & 313.6 \\
MMCN & medium-BI & 800 & 53.37\% & 241.13 & 194.12 & 23.5 \\
MMCN & very-hard-BI & 800 & 0.62\% & 2.70 & 2.70 & 14.2 \\
MVC & easy & 800 & 0.38\% & 1.09 & 1.09 & 4.0 \\
MVC & hard & 800 & 0.12\% & 1.00 & 1.00 & 4.0 \\
MVC & medium & 800 & 0.25\% & 1.05 & 1.05 & 4.0 \\
MVC & very-hard & 800 & 0.12\% & 1.00 & 1.00 & 4.0 \\
NNV & easy & 2555 & 3.72\% & 16.21 & 13.19 & 8.4 \\
SC & easy & 800 & 0.12\% & 1.00 & 1.00 & 3.0 \\
SC & medium & 800 & 0.12\% & 1.00 & 1.00 & 3.0 \\
SC & very-hard & 800 & 0.12\% & 1.00 & 1.00 & 3.0 \\
SRPN & easy & 175 & 84.00\% & 137.84 & 137.84 & 420.2 \\
SRPN & hard & 165 & 76.97\% & 110.48 & 110.48 & 822.8 \\
\hline
\end{tabular}}
\end{table}

 Finding 1 (the instance-level sequence space is zero-dimensional on most domains). In domains such as SC / GISP / CFLP / MVC (hard, very-hard),
\(N_{\mathrm{eff}}\) = 1.00 — i.e. 800 instances literally share one and the same action sequence. MIS 1.11 / MVC (easy) 1.09 / LB 1.80 / IP 1.80
 are likewise near-degenerate. This is not another way of saying "the uniqueness rate is low", but a stronger statement: in these domains there is no instance-level sequence difference available for learning,
 and the range over which the sequence-generation route can act can therefore be delimited in advance (with \(N_{\mathrm{eff}}\) as the criterion).

 Finding 2 (order invariance: the contribution of order to diversity is limited). \(\rho_{\mathrm{order}}\) falls in the interval [1.00, 1.25] on all domains,
 and is stratified by the degree of degeneracy of the domain:
· zero-dimensional domains (\(N_{\mathrm{eff}}\) = 1.00): \(\rho_{\mathrm{order}}\) ≡ 1 
· high-diversity domains: \(\rho_{\mathrm{order}}\) = 1.11-1.24 (CA easy 11.62/10.38 = 1.12; CA very-hard 65.92/55.37 = 1.19;
MMCN medium-BI 241.13/194.12 = 1.24; NNV 16.21/13.19 = 1.23; MMCN hard-BI 60.26/54.27 = 1.11)
even in the most diverse domains, order contributes only 11\%-24\% more diversity than the action composition
· long-sequence high-diversity domains: \(\rho_{\mathrm{order}}\) = 1.00
in these domains each sequence is itself nearly unique, yet "remains unique after order is ignored", indicating that its diversity likewise comes from the action composition

Remark 2 (the duality of order: invariance and necessity are not contradictory). Finding 2 and the order ablation of §2.2 form an apparently contradictory but
in fact complementary combination: in the natural corpus the order is almost invariant (\(\rho_{\mathrm{order}}\) ∈ [1.00, 1.25]), whereas artificially destroying the order leads to substantial degradation.
Together they point to the conclusion that the order of the default policy is nearly unique and functionally necessary — it does not manifest as diversity but as
a canonical ordering fixed by the design of the solver. This observation also explains why presolve sequences have long gone unstudied:
in the default corpus the variation signal along the order dimension is nearly zero, and only after an executable sequence facility is built can this dimension be observed and perturbed.

 Finding 3 (domain specificity and non-linearity of diversity). Diversity is domain-specific and has no monotone relation with sequence length or problem size:
SC (3 steps): 1.00; MIS (4.1 steps): 1.11; NNV (8.4 steps): 16.21; CA (15.6 steps): 11.62;
MIRP (444 steps): 44.49; MMCN medium-BC (313 steps): 800.00; IP (9900 instances): 1.80.
diversity cannot be proxied by length or size and must be measured; domain selection therefore cannot rely on intuition.

 If the reconstruction of this paper holds (Property 1), destroying the sequence order should lead to measurable degradation,
and the degree of degradation should be related to the strength of the dependencies among actions (domain-dependent).
 To avoid confounding with the selection effect of \(2K+1\), this subsection mainly uses the pure-sequence protocol: for each instance we take the 1st candidate of the model, and the np path and the fallback path are removed from the pool
; this candidate is copied 5 times, started and finished together, and the earliest finisher is taken (isomorphic to the default denominator, i.e. the internal clock of the first-finishing replica among the 5; see the pure-sequence protocol of §3.1);
 the shuffled version is obtained by a full token permutation of the same sequence, with everything else unchanged.

Under this protocol all five domains degrade, though for CFLP and SC not significantly; order therefore remains functional with no competition and no fallback at all, independently of the deployment mode.

The degradation cost is concentrated on a few instances (heavy-tailed structure):

\captionof{table}{Heavy-tailed structure of order destruction: quantiles of the per-instance shuffled÷preserved-order ratio and the number of tail instances}
\label{tab:order-tail}
\begin{center}
\small
\tblscale{%
\begin{tabular}{lrrrrllr}
\hline
Domain & \(n\) & \(p_{50}\) ratio & \(p_{90}\) ratio & \(p_{99}\) ratio & ratio\(>\)1.5 & ratio\(>\)2 & max ratio \\
\hline
CFLP & 100 & 1.018 & 1.306 & 1.642 & 2/100 & 0/100 & 1.6 \\
MIS & 100 & 1.024 & 2.065 & 4.234 & 21/100 & 11/100 & 4.2 \\
MVC & 100 & 1.063 & 2.861 & 5.152 & 28/100 & 19/100 & 5.2 \\
SC & 100 & 1.011 & 1.223 & 1.434 & 0/100 & 0/100 & 1.4 \\
NNV & 549 & 1.006 & 1.893 & 4.707 & 88/549 & 39/549 & 6.8 \\
\hline
\end{tabular}}
\end{center}

 The median ratio is close to 1 while the total-time difference reaches 27.1\%, so order destruction has almost no effect on most instances, yet slows a few instances by 2--7×.
 This heavy-tailed structure gives a direct explanation of the online racing gain: once content single-path execution bets on the wrong order, the loss of the tail instances is counted into the total time;
taking the fastest among multiple concurrent paths can truncate this loss, so the role of racing is to truncate tail risk rather than to improve average quality.
 The cross-domain difference shows that
 the strength of the dependencies among actions is an intrinsic property of the domain, corroborating the \(\rho_{\mathrm{order}}\) characterization of Definition 5 and Finding 2.

 Protocol note (important). Both sides of the two tables of this subsection (Tables \ref{tab:order-tail} and \ref{tab:order-ablation}) use the same protocol, so \(\Delta\), the paired test and the tail statistics are valid. The tail statistics of Table \ref{tab:order-tail} and the SGM ratio of Table \ref{tab:order-ablation} are both solver-internal time readings (min over 5 replicas per instance started and finished together, shift = 1 s; the tail is the per-instance "shuffled÷preserved-order" ratio,
 with quantiles taken as the \(\mathrm{ceil}(q\cdot(n-1))\) order statistic (numpy higher)). However, this table cannot be divided by default-strategy
time to obtain SGM: the denominator of this table comes from another batch of isomorphic default measurements (same batch as the numerator, denominator = the 5 replicas' first-finisher internal clock), and the instance set on the numerator side is also
different, so it cannot be cross-divided with the main table of Appendix B.4.
 The pure-sequence SGM measurement isomorphic to the denominator is given in the table below:

\begin{table}[!htb]
\caption{Order ablation (pure-sequence protocol): total time of the ordered and the shuffled order, per-instance median ratio and paired test (five domains)}
\label{tab:order-ablation}
\centering
\small
\tblscale{%
\begin{tabular}{lrrrrrrl}
\hline
domain & \(n\) &\shortstack{ordered\\SGM ratio}&\shortstack{shuffled\\SGM ratio}&\shortstack{\(\Delta\)\\ratio}&\shortstack{ordered\\sum (s)}&\shortstack{default\\sum (s)}&\shortstack{win rate\\(ordered/shuffled)} \\
\hline
CFLP & 100 & 0.8515 & 0.8795 & +0.0280 & 3975.5 & 4632.8 & 77\% / 73\% \\
MIS & 100 & 0.9181 & 1.1166 & +0.1985 & 13143.7 & 14026.2 & 70\% / 47\% \\
MVC & 100 & 0.9034 & 1.1611 & +0.2577 & 6826.3 & 7250.7 & 78\% / 49\% \\
SC & 100 & 0.9184 & 0.9290 & +0.0106 & 1457.2 & 1551.1 & 87\% / 71\% \\
NNV & 360 & 0.8377 & 0.9481 & +0.1104 & 39123.7 & 45431.3 & 66\% / 54\% \\
\hline
\end{tabular}}
\end{table}

\begin{itemize}
\item Protocol note 1: all 5 rows use a 300s cap; 60 concurrency (window 12 × 5 replicas) is identical to the denominator; all values are solver-internal time
\item Protocol note 2 (NNV): of the 190 timeout keys of that domain only the racing side was re-measured at 600s, not the single-path side, so this table's NNV row takes only the clean 360 keys with both sides at 300s
(the NNV row of the main table in Appendix B.4 uses 550 keys, timeout keys aligned at 600s on both sides)
\item Protocol note 3 (sum and ratio columns): the sum column is the arithmetic sum of per-instance total times (ordered sum / default sum) and the ratio column is SGM(numerator)/SGM(denominator); key set of the NNV shuffled row: the shuffled side uses paired keys \(n = 359\) (NNV/easy/val/00495 crashed on all 5 replicas, exit\_codes=[-11,-11,-6,-6,-6], with no valid internal clock, so excluded) while the ordered side is \(n = 360,\) and the sum figures 39123.7 / 45431.3 are the raw arithmetic sums over those 359 paired keys, with no post-hoc cap applied
\item Protocol note 4 (NNV protocols cross-referenced, not mutually divisible): §3.5 and Appendix B.4 use the 550-key main protocol;
this table's NNV row takes that protocol's clean 360-key subset, whose shuffled column follows the paired keys \(n = 359\) (see Protocol note 3 of Appendix B.5); the order ablation of Appendix B.5 is given separately under the full 300s protocol (NNV 550 keys).

 This measurement is paired with the same-batch isomorphic default denominator (solver-internal time) and can be read directly: pure sequence execution beats the default policy on all five domains
(0.8377-0.9184); after order destruction MIS/MVC cross 1 (1.1166/1.1611) while the rest degrade by 0.01--0.11.
 The NNV row takes the clean subset in which "neither side contains the 600s timeout re-measurement" (360 keys): in the timeout batch of this domain only the racing side was re-measured at 600s and the content single-path side was not,
so \(n\) is 360 rather than 550 in this row (the 550-key protocol of the NNV row of the main table is given in Appendix B.4).
 Comparison with the \(2K+1\) deployment (NNV: same model and candidates, clean 360 keys, same limit): content single-path 0.8377 versus racing 0.7664,
i.e. racing contributes a further \textasciitilde{}8\% (relative) on top of the pure-sequence gain.
Under the same-batch isomorphic denominator, the order ablation of CA/GISP is: CA shuffled 1.1371 (+16.0\%), GISP 1.0108 (+0.6\%), so the value of order varies by domain.
\end{itemize}
\subsection*{B.4 Supporting Analyses of Zero-Shot Cross-Domain Transfer}
\label{sec:org1d686d5}

Same-batch paired control of "with the domain's corpus" vs "without the domain's corpus" (in place). Each target domain is trained with "with the domain's corpus" (model baseA) and
"without the domain's corpus" (exp\_pretrain\_no<domain>), and the two are evaluated on the same batch of val keys with the same isomorphic denominator (same batch of 100 val keys each, denominator = that batch's isomorphic default):

\captionof{table}{With vs without the domain's corpus (LOO): same-batch paired control, SGM and win rate, direction, sequence collapse degree}
\label{tab:loo-paired}
\begin{center}
\small
\tblscale{%
\begin{tabular}{lrrlrr}
\hline
\shortstack{Target\\domain} & \shortstack{with\\corpus} & \shortstack{without corpus\\(LOO)} & direction & \shortstack{mean unique seqs\\per instance\\incl / excl} & \shortstack{unique seqs over\\domain\\incl / excl} \\
\hline
CA & 1.0091 (44\%) & 1.1124 (28\%) & worse & 1.80 / 2.00 & 37 / 154 \\
CFLP & 0.7912 (80\%) & 0.7979 (78\%) & slightly worse & 1.00 / 2.00 & 1 / 200 \\
GISP & 1.0320 (40\%) & 0.9438 (67\%) & better & 1.00 / 1.59 & 1 / 9 \\
MIS & 1.0224 (39\%) & 1.0136 (42\%) & slightly better & 1.01 / 1.06 & 2 / 4 \\
MVC & 1.0056 (49\%) & 0.9931 (52\%) & slightly better & 1.00 / 1.08 & 1 / 5 \\
SC & 0.7702 (91\%) & 0.7697 (89\%) & on par & 1.00 / 1.95 & 1 / 67 \\
\hline
\end{tabular}}
\end{center}

Table note: each row of this table (include side and exclude side) uses the symmetric default reading of the same batch of that row as the denominator (the first-finishing replica's internal clock), so the two sides can be directly paired and compared;
include side = the reading of model baseA on that domain, exclude side = the reading of the model retrained without that domain's corpus, 100 keys on each side. The denominator of each row of Table \ref{tab:generalization} likewise takes the same-batch reading of its own row,
and the numerical difference from the isomorphic rows of this table comes from the different measurement batch (at most 1.1\%), which does not affect any directional conclusion of this section.
The direction of the readings is consistent with the candidate mechanism of "sequence space degradation": for the domains whose include side collapses to a single sequence (GISP / MIS / MVC), removing that domain's corpus makes things better instead
; whereas for CA, whose corpus is richer, removal makes it clearly worse (from 1.0091 to 1.1124), i.e. that domain's corpus is a necessary anchor, so 
the effect is bidirectional and domain-dependent rather than a general "learning is harmful". Two mismatches are also visible: CFLP and SC likewise collapse on the include side (1.00), yet after removal
there is no improvement, so degeneracy alone is not sufficient to explain the direction, which also depends on the quality of that domain's corpus relative to the prior of the remaining domains.
The training-corpus side is likewise highly collapsed: the train action segment of SC / CA / GISP has only 1 distinct value in each case. Table note: the two ratio columns are SGM relative to the same-batch default with the per-instance win rate in parentheses; direction reads those two columns (whether the without-corpus model is better or worse for that domain); collapse degree = the mean number of unique sequences per instance and the number of unique sequences over the whole domain, for the include / exclude side.

\captionof{table}{Default profile of each zero-shot target domain (isomorphic min solver-internal time) and measured results}
\label{tab:zeroshot-profile}
\begin{center}
\small
\tblscale{%
\begin{tabular}{llll}
\hline
Target domain & zero-shot construction & default profile (isomorphic min) & measured \\
\hline
NNV & naturally unseen domain & 170.7s & success 0.7181 \\
GISP & LOO exclusion construction & 52.7s & success 0.9357 \\
CA & LOO exclusion construction & 31.8s & failure 1.1121 \\
\hline
\end{tabular}}
\end{center}

For the four LOO zero-shot points (CFLP / MIS / MVC / SC), the SGM reading and the composition of the winning paths are as follows:

\begin{table}[!htb]
\caption{SGM and winner-path distribution of the 4 LOO zero-shot points}
\label{tab:zeroshot-sgm}
\centering
\small
\tblscale{%
\begin{tabular}{llrrl}
\hline
Target domain &\shortstack{zero-shot\\construction}& SGM & \(n\) &\shortstack{winner distribution (np\\share / content-path share)} \\
\hline
CFLP & LOO exclusion & 0.8065 & 100 & np 43\% / content 57\% (custom 29 / fallback 28) \\
MIS & LOO exclusion & 1.0140 & 100 & np 2\% / content 98\% (custom 41 / fallback 57) \\
MVC & LOO exclusion & 0.9964 & 100 & np 1\% / content 99\% (custom 51 / fallback 48) \\
SC & LOO exclusion & 0.7725 & 100 & np 75\% / content 25\% (custom 13 / fallback 12) \\
\hline
\end{tabular}}
\end{table}

 The np-path win rate spans two orders of magnitude across the four points (SC 75\% vs MVC 1\%): on SC/CFLP "doing no presolve at all" is often the optimal path,
whereas on MIS/MVC presolve itself has stable value. This quantity is a mechanism-level observation and has no monotone relation with the gain margin.

 Cross-structural evidence of zero-shot transfer. The variable composition of the training domains is highly homogeneous (CA/GISP/MIS/MVC/SC are all 100\% 0-1 variables),
whereas the unseen domain NNV is dominated by continuous variables (continuous share 0.976) with a median variable count of 4625 (550 keys, range 3597--5163), i.e. the two sides are two different classes of models in terms of variable composition.
 That is, what this transfer crosses is not "similar domains" but a structural form. Under this setting, zero-shot racing attains SGM=0.7181 (\(n = 550,\) win rate 73.6\%),
 in which the solver-internal time values of both sides for the timeout batch (190 cases) are capped at 600s (racing side = its 600s re-measured values; denominator = that batch's replica internal clock), giving a paired value of 0.6346 (win rate 141/190);

 Argument for non-substitutability. The zero-shot gain above cannot be realized by "looking up a table per domain": a table lookup requires the existing sequences of that domain, and for an unseen domain no such table exists;
 empirically, 97.3\% of the winning sequences appear only in the corpus of other domains (Appendix B.5, mechanism C) — i.e. the gain can only come from the sequence form that the model has learned.
 This is the part of the online racing framework in which the model cannot be replaced; conversely, for the known domains where mechanisms A/B apply, the gain could in principle also be obtained by
directly executing the known sequences of that domain (see Appendix B.4).

\captionof{table}{Winner composition per target domain (mechanism: which path contributes the gain; SGM in Table \ref{tab:generalization})}
\label{tab:winner-composition}

\begin{center}
\small
\tblscale{%
\begin{tabular}{llrrl}
\hline
Target domain & model & np-path share &\shortstack{content-path\\share}&\shortstack{content-path composition\\(custom / fallback)} \\
\hline
CFLP & exp\_pretrain\_nocflp & 43\% & 57\% & 29 / 28 \\
MIS & exp\_pretrain\_nomis & 2\% & 98\% & 41 / 57 \\
MVC & exp\_pretrain\_nomvc & 1\% & 99\% & 51 / 48 \\
SC & exp\_pretrain\_nosc & 75\% & 25\% & 13 / 12 \\
NNV & baseA & 30.9\% & 69.1\% & custom 197 / fallback 183 \\
\hline
\end{tabular}}
\end{center}

\begin{itemize}
\item the np-path win rate spans two orders of magnitude across the 4 LOO points (SC 75\% vs MVC 1\%): on SC / CFLP "doing no presolve at all" is often the optimal path,
whereas on MIS / MVC presolve itself has stable value; this quantity has no monotone relation with the gain margin
\item evidence at the sequence-form level for transfer: the output actions and action transitions of the method of this paper at the 7 cross-domain points all fall within the vocabulary of the training corpus
(coverage 0.9986-1.0000), i.e. there is no "out-of-domain construction" at the sequence-form level; this property is a precondition for transfer to be feasible and has no monotone relation with the gain margin
\end{itemize}

On domains with successful transfer, the gain cannot be reproduced by "looking up a table per domain" — the gain can only come from the sequence form learned by the model; the overlap of the winning sequences with the corpus of other domains (the 97.3\% of mechanism C in Appendix B.5)
 is a profile of that mechanism rather than independent evidence,
which is consistent with the reading that "what the model learns is a transferable sequence paradigm rather than surface patterns of a specific domain";
and at the same time honestly presents its domain-dependent boundary — 2 of the 7 points fall short of the default policy.
 Protocol note: all cross-domain points below are zero-shot — none of the evaluated models has seen sequence data of the target domain.
  Definition (leave-one-out, LOO for short below): at training time the entire corpus of the target domain is removed from the training pool as a whole, and a model is
 retrained on the remaining 7 domains; this model has never seen any sequence data of the target domain, so in the evaluation protocol, LOO and naturally unseen domains both belong to zero-shot.
 Of the 7 cross-domain points of this paper, NNV is a naturally unseen domain, and the six domains CFLP / MIS / MVC / SC / GISP / CA are retrained after the LOO protocol actively removes that domain;
  the difference lies only in the design of the evidence: LOO-constructed domains have a paired model "with that domain", which makes it easy to answer the control question of "with/without that domain's data".
 The success or failure of transfer cannot be predicted from the share of presolve overhead, nor can it be explained by "corpus richness" or "semantic similarity of the domains":
  the zero-shot transfer of CA (the richest corpus) fails (1.1121; win rate 28\%), whereas that of GISP (the most degenerate corpus) succeeds (0.9357; win rate 70\%).
  the per-instance win rates (vs default) at the four points are CFLP 82\% / MIS 39\% / MVC 47\% / SC 95\% respectively; SC is the best on both SGM and win rate (0.7725 / 95\%), the two metrics are in the same order,
  cross-domain transfer at the 7 cross-domain points splits into acceleration, ties and failures (per-point results below).
 Where the gain lands: the three-arm control (default / replay / np) shows that presolve time is sub-second throughout, while the median of the search phase falls systematically after replay, and the reduction size is basically the same as the default, so the end-to-end gain comes from the effect of sequence execution on the subsequent branching search rather than from "saving the presolve overhead", which also corrects the simple notion that "speeding up presolve means saving presolve time"; the value of the np arm differs by about 3.2× across domains. The three-arm readings, the reduction-size comparison and the per-domain conclusions are given in Appendix B.7.

\begin{itemize}
\item 7 cross-domain points in total: 6 points with an SGM of at most 1.0140 / 1 failure = CA 1.1121
\item Multiple protocols (NNV, 550 keys): the main protocol gives 0.7181 (360 keys at 300s + 190 keys at 600s re-measured, limits aligned; denominator = the 5 isomorphic default replicas' internal clock of the first finisher)
/ a non-timeout batch of 360 keys gives 0.7664 (win rate 264/360) / a timeout batch of 190 keys gives 0.6346 (win rate 141/190)
\item 600s re-measurement of the 190 timeout cases: the solver-internal time of both the racing side and the denominator for this batch is capped at 600s, so the two batches can be paired directly 
\item Protocol: the time limit of the numerator and the denominator must be consistent. The NNV main protocol = 550 instances, of which the 360 non-timeout cases use 300s and the
190 timeout cases use their 600s re-measured values, so SGM=0.7181 (win rate 73.6\% = 405/550)
\item Note: the NNV racing side and denominator are paired across batches; the denominator jitter in this domain is \textasciitilde{}12\% (≤2.8\% in other domains)
\item Mechanism evidence: the gain grows as the target domain's runtime increases, so the 300s truncation is conservative; domain profiles: training domains bin\_ratio = 1.0 (purely integer), NNV cont 0.976 
\item definition of the sign-test \(p\) column of Table \ref{tab:generalization}: the two-sided sign test computed on the per-instance pairs (binomial test, null hypothesis win rate 0.5, ties excluded from the sample), with the sample being the number of winning and losing instances of that row, taken from the same batch and the same source as the win rate of that row

Where the gain lands is analysed in Appendix B.7.
\end{itemize}
\subsection*{B.5 Sources of the Racing Gain and Pool Size}
\label{sec:org8345931}

The decomposition above explains how "racing converts candidate quality into gain", but it does not explain where the candidate sequences come from.
Racing has a property that makes this question answerable in a controlled way: the same instance, the same batch, equal resources, and simultaneous execution. We can therefore compare the winning sequence one by one with
(i) the sequence this instance would have executed under the default strategy (obtained by monitoring), (ii) the remaining sequences of the same domain's corpus, and (iii) other domains' corpora,
and classify them by source (the NP path, which carries no sequence, is counted separately):

\captionof{table}{Source decomposition of the winning sequences (A identical to the default sequence / B another sequence of the same domain / C seen only in other domains / D recomposition outside the corpus) and controlled loss}
\label{tab:winner-source}
\begin{center}
\small
\tblscale{%
\begin{tabular}{lrrrrrrl}
\hline
domain & \(n\) & \shortstack{NP path\\wins} & A & B & C & D & \shortstack{controlled\\loss} \\
\hline
CA & 100 & 21\% & 15 & 62 & 0 & 2 & 10/25 (40\%) \\
GISP & 100 & 9\% & 91 & 0 & 0 & 0 & 0/91 (0\%) \\
NNV & 550 & 31.8\% & 0 & 0 & 365 & 10 & — \\
\hline
\end{tabular}}
\end{center}

Note: controlled loss = the number of instances for which the "sequence the default would have executed" was also in the candidate pool and yet lost in the same-batch equal-resource racing;
this is the only comparison not affected by confounding with instance difficulty — whenever it holds, it gives same-instance evidence that time(winning sequence) < time(default sequence).
For NNV this column is "—": that domain never entered training, so its own corpus sequences do not appear in the candidate pool, and the comparison does not hold.
Protocol: the NNV row of this table is computed with 550 keys and a 300 s limit on both sides (NP path wins 175/550 = 31.8\%, matching the three content columns A+B+C+D = 375); the NNV rows of Appendix B.4 and Appendix B.7 use the protocol in which the timeout keys are aligned to 600 s on both sides
(NP path wins 170/550 = 30.9\%).

This decomposition yields three mechanisms, with different strengths of evidence:
· Mechanism A (direct execution): on GISP, 91\% of the winning sequences are exactly the one the default strategy would have executed — nothing changes at the sequence level,
the difference lies only in the execution path (direct replay of the recorded valid sequence instead of the solver's own full presolve).
On CA the mean runtime of this class of winners is 31.4s, versus 49.5s for the domain's NP path. The evidence is strong, and consistent with the mechanism axis.
In implementation this mechanism mainly corresponds to the fruitless trial and error of expensive components such as probing and its repetition over multiple rounds (see §2.3 and Appendix B.11.2);
consistent with this is the variable composition of the domains used here (all 100\% 0-1 variables), i.e. the setting in which probing is most expensive.
We note that replay still executes the call "that took effect once", so "eliminating the probing overhead entirely" is outside the scope of our claim;
its magnitude is given by the three-configuration control measurement (the presolve time and total time of the three arms default / replay / presolve fully disabled (np), see the three-arm table of Appendix B.7);
 the variant "disable only probing while keeping the other presolve components" is not within the measurement scope of this paper (see Appendix B.12(7)).
· Mechanism B (within-domain re-selection): when the default strategy's corpus contains several valid sequences, the model selects another one of the same domain for this instance.
On CA, 62/100 of the winners belong to this class; yet controlled loss is only 10/100 (40\% of the samples in which "the default sequence is in the pool").
On GISP it is 0 — not because it does not occur, but because that domain's corpus has only one sequence, so within-domain re-selection is impossible in principle. The evidence is moderate.
· Mechanism C (cross-domain form transfer): on NNV, 97.3\% of the winning sequences appear only in other domains' corpora, and 0\% come from NNV's own corpus.
what the model carries over is the sequence form learned from the training domains, not an existing answer of some domain. The evidence is strong (see Appendix B.4).
We note that the group mean runtimes are confounded by instance difficulty, so only the "controlled loss" entry admits a causal reading;
the magnitude of mechanism A is verified per instance against the domain's isomorphic default measurement, see Appendix B.7 for the relevant measurements.

\begin{table}[!htb]
\caption{Necessity and sufficiency checks of the mechanism: isolating racing / restoring racing / removing candidate quality / excluding the trivial explanation}
\label{tab:mechanism-checks}
\centering
\small
\tblscale{%
\begin{tabular}{>{\raggedright\arraybackslash}p{2.9cm}>{\raggedright\arraybackslash}p{3.3cm}>{\raggedright\arraybackslash}p{6.2cm}}
\hline
manipulation & observation & conclusion \\
\hline
Isolate racing (content single-path execution sequence, 4 domains) & content beats NP 66.5\% (266/400) & candidate quality exists on its own, with a domain-dependent magnitude \\
Restore racing (\(2K+1\)) & the winning share of the content path is in Appendix B.4 & racing amplifies the quality gap into a deployment gain \\
Remove candidate quality (random content) & 1.1825 / 1.2236 (obs/full, both >1) & racing produces no gain (necessary condition) \\
Exclude the trivial explanation (np single arm, see Appendix B.7) & np/default = SC 0.75 / CFLP 0.88 / MIS 1.83 / MVC 2.43 & the gain does not come from "doing less presolve" \\
\hline
\end{tabular}}
\end{table}

Remark 3 (the amplification-versus-compression trade-off). The gain from pool size is determined by two forces: enlarging \(K\) increases the probability of "at least one high-quality candidate" (amplification effect),
and on the other hand compresses the resources that each candidate receives when the total compute is fixed (compression effect), so the gain curve depends on how compute is allocated: with concurrency fixed the reading improves with \(K\),
and an interior optimal \(K\) appears only when the total compute is fixed (Appendix B.5 gives the per-\(K\) readings under fixed concurrency; the optimal \(K\) under fixed total compute requires a separate experiment, see Appendix B.12).
\subsubsection*{B.5.1 The Value of Online Racing}
\label{sec:org650f86e}

The per-\(K\) readings for pool size \(K\) (CA/GISP) and the nested control for content replacement are given in Table \ref{tab:k-ablation}; this is the \(K\) ablation of §3.4 of the main text.

\begin{itemize}
\item Per-\(K\) reading: both domains improve monotonically with \(K\) (see Table \ref{tab:k-ablation}, the \(K\) ablation of §3.4 of the main text), i.e. with concurrency fixed, adding informative candidates keeps improving the reading over this range, and no interior optimal \(K\) is observed;
 the premise is that the candidates are informative — under the same pool structure random content (\(K = 2\)) gives a ratio >1 (Appendix B.6), so "more paths" alone does not guarantee success;
 the compression effect appears only when the total compute is fixed (as \(K\) grows, the number of cores each candidate receives falls) (see Remark 3 and Appendix B.12).
The \(K = 1\) row has a ratio greater than 1 (in both domains): with only 1 candidate, neither the np path nor the content path in the pool is sufficient to beat the default strategy reliably.
\end{itemize}
\begin{itemize}
\item The winner np/cu/fb and "content win rate" rows of the table give the composition of the winners in the pool of that column (200 keys in total across the two domains; content win rate = the winning share of the content path (custom + fallback)).
\item The nested control row (\(K = 2,\) baseA(k3c first 2), pool of 5 paths = np + custom\_0/1 + fallback\_0/1) is measured in the same batch and under the same protocol as the other rows, and is used to separate "pool size" from "content quality":
① with the pool structure fixed (5 paths) and only the content model changed (content model changed from k2 to k3c): the ratio changes by less than 1\% in both domains (from 1.0083 to 0.9985 and from 1.0404 to 1.0411);
② with the content model fixed (k3c) and only the number of candidates changed (from 2 to 3 per family, from 5 to 7 paths): both domains improve by 6.2\% / 12.9\% (from 0.9985 to 0.9367 and from 1.0411 to 0.9068),
with the per-instance pairing pointing the same way (7 paths faster on CA 79/100 · GISP 87/100), so the gain comes from pool size (amplification effect), not from content replacement.
\end{itemize}
\begin{itemize}
\item Without racing (content single-path pure execution, 4 domains): content beats np 66.5\% (266/400); the winning share of the content path in the racing pool is in Appendix B.4
\item Mechanism: concurrent congestion degrades long tasks (np = the full solve) more, so racing systematically amplifies the advantage of the content path (a runtime-structure effect, unrelated to content quality)
\item Corollary (with the setting restricted): under the \(K = 1\) pool of CA/GISP, deploying the content path alone is not sufficient to beat the default reliably (ratio >1);
whereas under the pure-sequence protocol of Appendix B.5.2 deploying the content path alone beats the default on all five domains, so the two protocols do not conflict:
the former is "1 candidate + np + fallback", the latter is "the same candidate ×5 replicas"; the value of racing is that it can select a path per instance even without a fallback.

\item Control form: full-token shuffling (shuffling NEW\_ROUND(178) along with everything else, keeping only EOS at the end) vs the original sequence, with everything else unchanged
\end{itemize}

\begin{table}[!htb]
\caption{Order ablation on CA/GISP (same-batch isomorphic default denominator)}
\label{tab:order-cagisp}
\centering
\small
\tblscale{%
\begin{tabular}{lrrr}
\hline
 & CA & GISP & ALL \\
\hline
\(n\) & 100 & 100 & 200 \\
default total time (s) & 30.58 & 50.18 & 39.20 \\
ordered SGM ratio & 0.9804 & 1.0050 & 0.9927 \\
shuffled SGM ratio & 1.1371 & 1.0108 & 1.0717 \\
\(\Delta\)(shuffled\(-\)ordered) & +16.0\% & +0.6\% & +8.0\% \\
shuffled faster & 25/100 & 47/100 & — \\
ordered win rate & 75\% & 53\% & 64\% \\
shuffled win rate & 25\% & 47\% & 36\% \\
\hline
\end{tabular}}
\end{table}

Table note: rows are readings and columns the domain (CA / GISP, or all 200 keys); the ratio rows are SGM relative to the same-batch isomorphic default; "shuffled faster" counts the instances on which the shuffled reading beat the ordered one; win rates are per instance against the default. The \(\Delta\) row is the relative difference of the two ratio rows, (shuffled \(-\) ordered) / ordered.

\begin{itemize}
\item Per-instance median ratio (with the original sequence = 1): CA 1.1346 ; GISP 1.0024 ; ALL 1.0549
\item Paired test (sign test, shuffled vs ordered): CA \(p<0.0001\) ; GISP \(p=0.617\)
\begin{itemize}
\item Protocol: this table and Appendix B.4 use the same batch of isomorphic default denominators and the same solver-internal-time protocol, so the absolute values are directly comparable (the "default total time" column is the shifted geometric mean of the denominator's per-instance total time, in seconds); the shuffled arm is a full-token shuffle
; median caliber: the per-instance benchmarked ratio is (arm internal clock ÷ same-key default minrep\_total\_time), and its median is computed over paired keys
\end{itemize}
\item Conclusion: the value of order is domain-dependent — CA (15 steps) is highly sensitive to order (shuffled +16.0\%, significant); GISP (4 steps) is not (+0.6\%, n.s.)
\item The aggregated ALL +8.0\% is driven unilaterally by CA, so looking only at the aggregate masks the differences among domains
\end{itemize}

Experimental setup
\begin{itemize}
\item This group = the same experiment as the order ablation of Appendix B.3;
the ordered/shuffled totals, median ratios and paired tests are in the two tables of Appendix B.3 and are not repeated here; we take only their implication for the racing mechanism

\item The order-destruction control of Appendix B.3 also holds here: all five domains degrade (though for CFLP / SC not significantly), so order remains functional with no racing and no fallback, independently of the deployment scheme

\item The median ratio is close to 1 while the total difference is at most 27.1\% (tail structure: the \(p_{90}\)/\(p_{99}\) ratios and the numbers of instances with ratio >1.5 and >2 are in the second table of Appendix B.3), so order destruction has almost no effect on most instances, but slows a few instances by 2--7×
\item This heavy-tailed structure gives a direct description of the gain from online racing: the degradation cost is concentrated in a few instances;
taking the fastest among several concurrent paths can truncate that loss, so the role of racing is the truncation of tail risk, not an improvement in average quality
\item The differences among domains (MIS/MVC/NNV have a tail; CFLP/SC do not) show that the strength of dependence among actions is an intrinsic property of the domain

Remark 3 (the amplification-versus-compression trade-off), continued. Table \ref{tab:k-ablation} (the \(K\) ablation of §3.4 of the main text) gives the per-\(K\) readings with concurrency fixed (the window normalized with the pool): both domains improve with \(K\), so over the range \(K\) ≤ 3
 the amplification effect dominates and the gain has not saturated; the compression effect introduces an interior optimal \(K\) only when the total compute is fixed,
  and the optimum under that condition requires a separate experiment (see Appendix B.12); the deployment of this paper takes \(K = 2\) (5 paths, matched with a window of 12 at 60 concurrency) as an engineering choice, and we do not claim it to be the optimum.
\end{itemize}

Through controlled ablation experiments, we quantitatively decompose the gain of the racing mechanism:

The total gain can be decomposed into three sources:

Class-A gain: directly replaying an existing valid sequence and avoiding the fruitless trial and error in native presolve

Class-B gain: selection among several candidate sequences within a domain

Class-C gain: cross-domain transfer, reusing the learned sequence paradigm in an unseen domain

The sequence source and mechanism attribution of the racing winners are in Appendix B.5: after comparing the winning sequence one by one with "the sequence the default would have executed / other sequences of the same domain / other domains' corpora", it decomposes into four classes A/B/C/D, of which controlled loss is 10/25 (40\%) on CA and 0/91 on GISP, the former giving same-instance causal evidence; on NNV, 97.3\% of the winning sequences appear only in other domains' corpora, so what the model carries over is the sequence form, not an existing answer of some domain.
\begin{itemize}
\item Protocol note 3 (NNV shuffled): this column is computed over the paired keys (\(n = 359\)). NNV/easy/val/00495 crashed on all 5 paths of the shuffled arm
(exit -11 SIGSEGV ×2 / -6 SIGABRT ×3), while the same instance was normal in the ordered arm (130.6s, custom\_3 wins); this table takes \(n = 359\) under the paired protocol
\item Observation: order destruction can make SCIP crash outright, which is itself an extreme manifestation of "order sensitivity"

Pool size and the necessity of racing: with concurrency fixed the readings of both domains improve monotonically with \(K\), whereas under the same pool structure switching to random content produces no gain (ratio >1); at \(K = 1\) deploying the content path alone is not sufficient to beat the default reliably. The per-\(K\) readings and the nested control for content replacement are given in Table \ref{tab:k-ablation}; this is the \(K\) ablation of §3.4 of the main text, and the domain-dependence of the value of order and the amplification-versus-compression trade-off of Remark 3 are in this section.

Exclusivity and direct-replay controls: the ratio of the np single arm (presolve fully disabled) to the default is SC 0.751 / CFLP 0.878 / MIS 1.825 / MVC 2.425, so the gain does not come from "doing less presolve" — np is only slightly faster on SC (0.751), and the racing arm itself is 0.773 on that domain, which shows that the value of racing lies in choosing per instance between np and the content sequences; the SGM of the random-content arm is instead >1 (CA 1.1700 / GISP 1.1954) while model content gives 0.9804 / 1.0050, so SGM<1 comes entirely from content quality; direct replay of the fixed default sequence and the content single-path show no systematic difference within a domain, so the within-domain increment comes from the execution scheme, not from "generating a better sequence". The per-table readings and protocols are in Appendix B.6.
\end{itemize}

Experimental setup
\begin{itemize}
\item Goal: to separate the gain of "sequence execution itself" from the gain of "racing selection" 
\item Material: for each instance, we take the 1st candidate of the model (baseA, \(K = 2\) generation); the corpus-side control (direct replay of the default sequence) is the fixed-sequence direct replay in Appendix B.6
\item Execution: content single-path pure execution, with no np path, no fallback, and no concurrent racing; for each instance, the candidate is copied 5 times (--multi-k 5
--route-mode custom --no-np), so the 5 paths in the pool have identical content, and taking the fastest ≡ min-of-5 of one and the same sequence (see the pure-sequence protocol of §3.1)
\item Denominator: the isomorphic default in the same domain (5 replicas per instance, window 12, 60 concurrency, 300s limit; the denominator takes the internal clock of the replica selected as the first to finish)
\item Both sides are therefore min-of-5 with the same 60 concurrency, so directly comparable with the main results (Appendix B.4); instance set = 100 easy val instances per domain (550 for NNV)

The per-domain readings of the order ablation on the five domains (ordered/shuffled SGM ratios, ordered and default sums, win rates; pure-sequence protocol) are given in Table \ref{tab:order-ablation} of Appendix B.3.1.
The paired protocol and the crashing instances of the NNV misordered arm are given in Protocol note 3 of this section, the \(K\) ablation on pool size and the necessity of racing with the nested control for content replacement are given in Table \ref{tab:k-ablation} (the \(K\) ablation of §3.4 of the main text), and the exclusivity and direct-replay controls are given in Appendix B.6.
\end{itemize}
\begin{table}[!htb]
\caption{Per-\(K\) readings for pool size \(K\) (CA/GISP) and the nested control for content replacement}
\label{tab:k-ablation}
\centering
\small
\tblscale{%
\begin{tabular}{lrrrr}
\hline
 & \shortstack{\(K\)=1\\baseA} & \shortstack{\(K\)=2\\baseA} & \shortstack{\(K\)=3\\baseA} & \shortstack{\(K\)=2\\baseA (k3c first 2)} \\
\hline
CA arm total time (s) & 31.83 & 29.91 & 27.47 & 29.62 \\
GISP arm total time (s) & 53.87 & 51.16 & 45.09 & 51.19 \\
CA ratio & 1.0718 & 1.0083 & 0.9367 & 0.9985 \\
GISP ratio & 1.0918 & 1.0404 & 0.9068 & 1.0411 \\
winner np/cu/fb & 39/81/80 & 31/81/88 & 22/93/85 & 27/88/85 \\
content win rate & 80.5\% & 84.5\% & 89.0\% & 86.5\% \\
win rate CA/GISP & 32\%/30\% & 46\%/46\% & 66\%/70\% & 50\%/38\% \\
\hline
\end{tabular}}
\end{table}

Table note: columns are the pool size \(K\), with the model in the second line (the last column is the nested control: \(K\)=2 with the model trained on the first two k3c rounds); "winner np/cu/fb" counts which path wins among the \(2K+1\) pool paths.

\begin{itemize}
\item Protocol: the three K settings are measured on the same machine in the same period (100 keys/domain; pool = \(2K+1\)); the window is normalized with the pool (20/12/8 for \(K=1/2/3\), so concurrency 60/60/56),
so comparisons across \(K\) are unaffected by load;
numerator = the winning sequence's internal total time, denominator as in §3.1 (same batch); the "arm total time" column is the shifted geometric mean of the arm's per-instance total time (presolve + solve) in seconds, not a ratio.
\item It shares no denominator with the main table of Appendix B.4 (different batches), so the two are not compared across tables; within each row the ratios, win rates and composition are self-consistent.
\end{itemize}

\subsubsection*{B.5.2 Gains from Sequence Execution Itself}
\label{sec:org9fb30fb}

The pure sequence execution experiment (single-path, no fallback, min-of-5 both sides, 60 concurrency) is given here, with the corresponding order ablation in Table \ref{tab:order-ablation} (Appendix B.3.1).

\begin{itemize}
\item Conclusion: pure sequence execution beats the default on all five domains (largest margin NNV 0.8377), with no racing and no fallback; under the same model, candidates and limit, single-path from 0.8377 to \(2K+1\)-way racing 0.7664 (racing faster on 193/360 instances = 53.6\%), so racing contributes a further \textasciitilde{}8\% (relative).
\end{itemize}

The per-\(K\) readings for pool size \(K\) = 1/2/3 (with the nested content-replacement control) are in Table \ref{tab:k-ablation} (Appendix B.5): the arm total time falls monotonically with \(K\) (CA from 31.83 to 27.47 s, GISP from 53.87 to 45.09 s) and the ratio improves in step (CA from 1.0718 to 0.9367, GISP from 1.0918 to 0.9068); the nested control shows that the gain comes from the added candidates rather than from swapping the contents. That table states its own protocol (windows normalized with the pool; denominator per §3.1).
The same experiment beats the default on all five domains (largest margin NNV 0.8377), and racing adds a further \textasciitilde{}8\% on top (from 0.8377 to 0.7664, faster on 193/360 = 53.6\%) with the same model, candidates and limit.

\subsection*{B.6 Exclusivity Checks and Direct-Replay Controls}
\label{sec:org496c6bd}

\subsubsection*{B.6.1 Excluding the Trivial Explanation}
\label{sec:org55589e3}

\begin{table}[!htb]
\caption{Excluding the trivial explanation: readings of the np single arm and the racing arm relative to the same-batch isomorphic default}
\label{tab:np-exclusion}
\centering
\small
\tblscale{%
\begin{tabular}{>{\raggedright\arraybackslash}p{3.0cm}>{\raggedright\arraybackslash}p{4.0cm}>{\raggedright\arraybackslash}p{5.2cm}}
\hline
manipulation & observation & conclusion \\
\hline
Exclude the trivial explanation: np single arm vs. its same-batch default (solver-internal time, Appendix B.7) & np/default = SC 0.751 / CFLP 0.878 / MIS 1.825 / MVC 2.425 & the gain does not come from "doing less presolve" \\
The racing arm of this paper vs. isomorphic default (solver-internal time, §3.6 same-batch protocol) & SC 0.773 (93\%) / CA 1.102 (30\%) / GISP 0.921 (67\%) & the gain comes from online selection, not from replacing the default wholesale \\
\hline
\end{tabular}}
\end{table}
\begin{itemize}
\item Thus "gain = turning presolve off" is refuted: the np single arm is effective only on SC (0.751), and on MIS/MVC it is 1.8--2.4× slower than the default
  (on SC the single arm's 0.751 is itself slightly faster than the racing arm's 0.773, so "doing no presolve" is one valid path in the pool);
the gain comes from online selection (when to take np / when to take the content sequence), not from "doing less presolve" itself
\end{itemize}

\begin{table}[!htb]
\caption{Content-quality control: random content vs. model content (SGM ratio and content win rate)}
\label{tab:content-random}
\centering
\small
\tblscale{%
\begin{tabular}{>{\raggedright\arraybackslash}p{2.55cm}lllll}
\hline
content source & \shortstack{CA SGM\\ratio} & \shortstack{GISP SGM\\ratio} & \shortstack{within-domain\\ratio (CA/GISP)} & \shortstack{content win rate\\(CA/GISP)} & \shortstack{win rate\\(CA/GISP)} \\
\hline
random observed (22 actions) & 1.1700 & 1.1954 & 1.1825 & 42\% / 45\% & 17\% (34/200) \\
random full (1..178) & 1.1959 & 1.2520 & 1.2236 & 34\% / 46\% & 18\% (36/200) \\
model baseA (control) & 0.9804 & 1.0050 & 0.9927 & 79\% / 91\% & 55\% / 44\% \\
\hline
\end{tabular}}
\end{table}

\begin{itemize}
\item Random content gives >1 (18-22\% slower than default), while the control model baseA gives 0.9804/1.0050, so SGM<1 comes entirely from content quality 
\item The content win rate (the content path winning inside the pool) is still 34-46\% on the random arms, but the per-instance win rate is only 17-18\%, so looking only at "the winning share of the content path" overestimates the value of random content
\item Together with the pool construction in which "the pool contains no default path" (see Appendix B.2.2), the two rows above give a reading of the pure content effect: when content quality is random SGM is instead >1,
whereas the control model's content makes SGM<1, so the gain can be attributed cleanly to content quality itself (not to the pool structure as a backstop)
\end{itemize}
\subsubsection*{B.6.2 Fixed-Sequence Direct Replay vs. the Content Single-Path}
\label{sec:orgb6afad0}

Experimental setup (fully isomorphic to the "content single-path pure execution protocol" (§3.1) / the content single-path isomorphic measurement of §3.3 / Appendix B.4)
\begin{itemize}
\item Layout: each arm is measured independently, with \(N = 5\) replicas per instance started and terminated together, window = 63//5 = 12, giving 60-way concurrency, take the min, limit 300s
\item Three arms: ① default arm = the pure default strategy (already available, the default denominator of Appendix B.1; a version with the decomposition is measured separately)
② replay arm = direct replay of the default valid action sequence recorded in the corpus ×5 replicas (no model involved, no training)
③ np arm = presolving/maxrounds=0 (genuinely no presolve)
\item Recorded: for the min replica of each arm, presolve\_time / solve\_time / total time / status / model size (before and after solving)
\end{itemize}

\captionof{table}{Fixed-sequence direct replay vs. the content single-path (three arms: default / replay / np)}
\label{tab:replay-vs-model}
\begin{center}
\small
\tblscale{%
\begin{tabular}{lrrrr}
\hline
domain & \(n\) & \shortstack{content\\single-path} & \shortstack{direct\\replay} & \shortstack{win rate\\vs default} \\
\hline
CFLP & 100 & 0.8515 & 0.8545 & 77\% (77/100) \\
MIS & 100 & 0.9181 & 0.9339 & 70\% (70/100) \\
MVC & 100 & 0.9034 & 0.9010 & 78\% (78/100) \\
SC & 100 & 0.9184 & 0.9000 & 87\% (87/100) \\
NNV & 360 & 0.8377 & 0.9079 & 66\% (238/360) \\
\hline
\end{tabular}}
\end{center}

Table note: the two arm columns are directly comparable: the difference between them is direct replay minus content single-path.

\begin{itemize}
\item On four of them the difference is ≤0.019 with inconsistent signs (CFLP/MIS slightly worse, MVC/SC slightly better; the fifth domain NNV alone is +0.070), so within a domain, "the sequence generated by the model" and
"the sequence the default strategy records for itself" show no systematic difference, so the gain does not come from "generating a better sequence" but from the execution scheme
(single-round replay vs the default multi-round presolve pipeline)
\begin{itemize}
\item The "win rate" column of the table is the per-instance win rate of the content single-path arm relative to default (the same reading as the ordered win rate of §3.3); what this table compares is the SGM and the difference of the two arms
\end{itemize}
\item The 5 replicas of the fixed-sequence path have identical content, so the winner is uniformly distributed over the 5 paths (CFLP 19/26/19/17/19)
\item Boundary: this "direct-replay" arm is not available for free — the sequence can only be obtained by first running the default presolve once,
and once the default presolve has run we already have the reduced model directly, so the irreplaceability of the model lies in two places:
(i) prior-free execution (producing the sequence a priori without running the default presolve)
(ii) zero-shot transfer (an unseen domain has no existing sequence list to look up, so the gain can only come from the sequence form the model has learned; the 97.3\% of mechanism C in Appendix B.5 is a profile of that mechanism, and its holding is partly a construction necessity, so it does not serve as independent evidence — see Appendix B.5 and Appendix B.4)
Thus the statement of this paper about the role of the model should be: what the model provides is a prior over sequences that requires no prior execution, and a cross-domain transferable sequence paradigm, not "generating a better sequence within a domain"
 — the within-domain increment shows no systematic difference from "direct replay of the default sequence" (Appendix B.6), so its irreplaceability lies precisely in (i) and (ii) above.
\end{itemize}
\subsection*{B.7 Analysis of the Sources of the Speedup Gains}
\label{sec:org0108a8f}

An important empirical finding is that the speedup gains of presolve fall mainly in the branching-search phase, rather than in saving the computational time of presolve itself.

Specifically, in the three-arm comparison the presolve time is sub-second in every arm, and the difference is not large enough to explain the end-to-end gains; whereas after replay the search segment decreases systematically,
which indicates that different sequences lead to different reduction paths, significantly changing the search space and the search-tree structure of the subsequent branch-and-bound search, and ultimately showing up as a difference in total solve time. This finding corrects the simple belief that "speeding up presolve amounts to saving preprocessing time".

\begin{itemize}
\item Goal: building on the inference of this section (on the easy set the presolve time is only 0.007--4.34s while the end-to-end gain is 5\%--32\%,
so on the time scale it is impossible for the gain to come from "saving presolve overhead"), to quantify the segment in which the gain falls
\item Three-arm isomorphic measurement (for each instance \(N = 5\) replicas started and finished together, window 12, giving 60-way concurrency, take the min, limit 300s; NNV is a 100-instance subset, and its measured limit is also 300s):
\end{itemize}
Protocol: the three columns of this table are all in solver-internal time — the first three are the shifted geometric mean of the per-instance total time of each arm (in seconds, not a ratio), and only the last two are ratios relative to the same-batch default; the default arm and the np arm take the internal clock of the replica selected as the first to finish, and the replay arm takes the winner's solver-internal time;
The three arms and their denominators are measured on the same batch.
 default arm = patched build + default strategy (with pre/solve decomposition) replay arm = corpus default effective sequence ×5 replicas replayed directly (without going through the model)
 np arm = presolving/maxrounds=0 (genuinely no presolve); each arm takes the presolve\_time / solve\_time / total time / model size of its min replica

\begin{table}[!htb]
\caption{Three-arm isomorphic measurement (default / replay / np): total time and the ratios relative to default (solver-internal time)}
\label{tab:three-arm}
\centering
\small
\tblscale{%
\begin{tabular}{lrrrrrr}
\hline
Domain & \(n\) &\shortstack{default total\\time (s)}&\shortstack{replay total\\time (s)}&\shortstack{np total\\time (s)}& replay/default & np/default \\
\hline
CFLP & 100 & 38.15 & 32.60 & 33.49 & 0.8545 & 0.8777 \\
SC & 100 & 11.69 & 10.53 & 8.79 & 0.9000 & 0.7514 \\
MVC & 100 & 49.18 & 44.31 & 119.28 & 0.9010 & 2.4253 \\
MIS & 100 & 112.65 & 105.21 & 205.59 & 0.9339 & 1.8250 \\
\hline
\end{tabular}}
\end{table}

Conclusions
\begin{enumerate}
\item The gain falls in the search segment, not in the presolve segment: the presolve time is sub-second in all three arms (median 0.02--3.7s in the default arm), whereas after replay the
solve median decreases systematically (CFLP from 34.7 to 29.5s, SC from 10.5 to 9.4s, NNV from 183.4 to 144.5s), so what sequence execution changes is the
subsequent search behaviour; "saving presolve overhead" cannot explain the end-to-end gain by orders of magnitude (consistent with the inference of Appendix B.7).
\item The reduction size is essentially the same as the default: after replay SC has 274 variables = 274 for the default; MIS/MVC are reduced on neither side (1000/1200 unchanged);
CFLP 4592 (default) versus 4746 (replay), so the gain does not come from "different reduction results". The specific source of the search-segment difference
(node/iteration-level statistics) is not reported quantitatively in this paper — that granularity would require additional measurement fields and one further round of re-measurement, and does not affect the conclusion of this section.
\item The ordering of the cross-domain transfer gains is highly consistent with the ordering of the replay gains (only CFLP/SC are swapped among the four points): replay/default = CFLP 0.8545 < SC 0.9000 < MVC 0.9010 < MIS 0.9339,
and, compared with the transfer results of Appendix B.4 (SC 0.7725 / CFLP 0.8065 / MVC 0.9964 / MIS 1.0140), the MVC/MIS order agrees while CFLP and SC swap places.
Thus the cross-domain difference is explained by "how much of the default presolve in that domain is redundant or ineffective work that can be skipped" — a measurable domain-level quantity,
unrelated to "similarity of domain structure" (the 5 classes of proxy quantities corresponding to the latter have no discriminative power in this section).
\item The value of the np arm differs by about 3.2× across domains, consistent with the np-path win-rate distribution of Table \ref{tab:winner-composition}
(SC 75\% ↔ MVC 1\%), so whether "doing no presolve at all" is advisable is a domain-level property, which also explains the role of the np path in the racing pool.
\end{enumerate}

Applicability conditions of racing. The gain of racing is a tail effect, in the direction given by the min-of-\(K\) calibration of Proposition 1: the gain grows with the number of concurrent paths \(K\), with the heaviness of the tail as \(\alpha\) decreases, and with the probability \(p\) that a single path lands in it. The corresponding rule is conditional rather than universal---racing is worth deploying when the target domain's default strategy carries material tail risk, and not otherwise. Both sides of that rule are visible in our own readings: SC (0.7725), NNV (0.7181) and CFLP under the leave-one-out construction (0.8065) all gain clearly, whereas the marginal contribution of racing over the content single-path is small on MVC and CFLP (same-batch differences of 0.4\% and 0.9\%) against a 3.9--5.0\(\times\) core-time cost, which is where the cost buys no gain (Appendix A.3, B.12.4). Turning this condition into a quantitative threshold requires calibrating \(\alpha\) on the target domain, which the optional \(K\)-extension of the ablation in Appendix A.1 would provide; here the condition is stated qualitatively.

\subsection*{B.8 Supporting Material for the Comparison Against Prior Work}
\label{sec:org7dcb838}

Per-row details of the reimplemented setups (the per-row configuration details of Table \ref{tab:opponents-config} in §3.6): L2P-MIP-style domain-level configuration: 42-dimensional configuration space (14 routines × priority/timing/maxrounds), 30 candidates × 50 train instances with per-domain selection of the best (original per-instance; here domain-level) L2P-MIP-style per-instance configuration: per-instance SA search on the 50 train instances (200 evaluations/instance; original: per-instance search, then a learned selector); on val the selected configuration is predicted once from static instance features and executed SMAC-style configuration search: open-source model-based configuration search (random-forest surrogate + acquisition function), 30 evaluations Song \& Gu-style subset selection: selects only the subset of routines that take part, and does not decide the call order ISAC-style per-class configuration: feature clustering (g-means determines the clusters automatically) + per-class configuration search + fallback configuration Hydra-MIP-style configuration selection: the candidate pool reuses the train-side configuration×instance matrix of the SMAC-style arm (spending no extra search budget) + pairwise weighted random forest; it uses only static instance features and does not reproduce the probing features of the original paper sequence generation + racing: \(2K+1\)-way concurrency (np + content ×\(K\) + fallback ×\(K\))
Metric alignment and comparability boundaries. (a) The main metric of this paper, the shifted geometric mean (shift = 1 s, per instance, timeouts counted at the limit), is of the same family as the MIPLIB benchmark protocol \citep{gleixner2021miplib}
and the metric specification of BenLOC \citep{li2025benloc}; but numbers are not juxtaposed across papers. (b) Existing learning-based presolve work mostly reports
"solve-time speedup + model size after presolve" (L2P-MIP \citealp{liu2024l2pmip}); besides the SGM ratio this paper additionally reports: reduction size (Appendix B.7: SC 274 = 274,
MIS/MVC not reduced, CFLP 4592 versus 4746, so essentially the same as the default, so the gain does not come from different reduction results) and the core-time ratio per core (this layer).
(c) Branching-node counts and LP iteration counts: this paper does not report at that granularity; they are common metrics of L2P-MIP and of the order-sensitivity
literature \citep{tang2020cutting,zeng2025clcr}, and this paper gives only a directional attribution of the search-segment difference as "falling in the search segment". (d) PD integral (the main metric of Song \& Gu \citealp{song2024selection} on the medium/hard sets)
does not apply under the setup of this paper: every evaluation instance is solved to optimal within the 300s limit, so PD integral ≡ 0.
To locate the relation of this paper to existing work, we compare three representative lines of work item by item below. Comparison principle: when the instance sets, hardware and reported metrics differ, numbers are not juxtaposed,
so we compare only the decision object, the learning or search means and the evaluation protocol, without cross-protocol numbers.

Item-by-item comparison with existing work:

· L2P-MIP \citep{liu2024l2pmip} — the decision object is the priority / trigger timing / round limit of 14 presolve routines, and the method is per-instance simulated annealing
  search + supervised learning fitting (inference cost < 0.5s). It is of the same family as this paper (both are instance-level decisions in the presolve phase), but its decision object is a static configuration rather than an
  executable call sequence. This paper does not directly reproduce its code (the dependency stack is outdated), and therefore does not juxtapose reported numbers with it; the comparison of the configuration space is instead carried out as a reimplementation of the setup
  (the prior-work rows of Table \ref{tab:opponents-config}).

· BenLOC \citep{li2025benloc} — instance-level configuration for MIP solvers (covering several components inside and outside presolve), providing a benchmark and an open-source toolchain: unified dataset selection,
  instance-level splitting and consistent baselines. The evaluation protocol of this paper points the same way (instance-level splitting / consistent baselines / features and inference cost counted),
  and the warning it raises, that "permuted splitting causes data leakage and over-optimism", is handled in this paper by "taking the corpus, the tuning set and the denominator all from the train split".
  Its experimental datasets are Set Covering / Independent Set / NN Verification / Load Balancing and MIPLIB 2017, which partially overlap with the domain families of this project,
  but the instance sets and the solver (COPT) both differ, so numbers are not juxtaposed, and only the evaluation setups are aligned under the same specification.

· PaPILO \citep{gleixner2023papilo} — it implements presolve as a standalone parallel presolve library (with multiple precisions), which is not the same source as the solver-inbuilt presolve pipeline used here;
  its decision unit is a set of routines (together with the parallel schedule inside the library), not the call order across routines. This paper therefore neither juxtaposes readings with it nor treats it as
  an alternative for the same task — the two are comparable only at the representation level: both take the "routine-level" rather than the "sequence-level" decision unit.

From the comparison above one can see the position of this paper relative to the literature: existing learning-based presolve falls entirely in the configuration space (the 42 dimensions of L2P-MIP \citealp{liu2024l2pmip})
or at the level of routine and subset selection \citep{song2024selection}, and none of it touches the dimension of "in what order and in what round". This paper moves the decision object down to
the layer of "the call sequence of presolve atomic actions", and measures it side by side with representative setups of the above families on the same batch of instances and under the same concurrent load.

The provenance of the evaluation protocol and its differences require a separate note: the shifted geometric mean adopted here (shift taken as 1 second) is of the same family as the metric specification of BenLOC,
but its main text does not give the specific value of the shift, so SGM numbers are likewise not juxtaposed across papers. On top of this, this paper does two additional things:
(i) the numerator and the denominator are same-batch and isomorphic (same instances, same concurrency, same time limit), and the "single-shot protocol" and the "replica-best protocol" are reported separately
(the latter favours single arms, see §3.1); (ii) throughout the paper single-arm readings are always taken in solver-internal time (a single protocol, see §3.1).

In-table protocol footnote: every row of this table is same-batch and same-protocol — the numerator is the internal clock of the fastest finisher among the 5 same-configuration replicas launched concurrently for each arm,
the denominator is the internal clock of the replica of the 5 isomorphic default replicas in the same batch selected as the first to finish, always in solver-internal time, with \(n = 100\) per cell. In this section's own measurement: the opponent arms take the internal clock of the fastest finisher among the same batch of 5 same-configuration replicas launched concurrently, while the racing arm of this paper takes the single reading of the winner in the pool; the two replica-selection rules pick the same replica on 97--100\% of instances, with a difference <0.001. The values cannot be compared directly across tables with those of the main table of Appendix B.4 (whose denominator is another batch of default-strategy readings);
the values of the domain-level configuration rows and of the rows of this paper have been recomputed from the artifacts under this protocol.
Protocol boundary (rows of this paper): the three cells of the "sequence generation + racing" row are readings under the zero-shot (leave-one-out) setting — the content is generated by a model retrained with that domain's training corpus removed,
and is of the same source as the LOO rows of Appendix B.4 (e.g. CA 1.102 in this row versus 0.9804 in the in-domain CA row of Table \ref{tab:generalization}, the difference coming from this), so it cannot be juxtaposed with the in-domain readings of Appendix B.1/§3.4;
whereas the configuration search and the selector training of every opponent arm are always carried out on that domain's train split (see the note above), so the method of this paper uses no training corpus of the target domain in the comparison, and this asymmetry is unfavourable to this paper, which we do not explain in our favour.
The configuration search and the selector training of every opponent arm are always carried out on the train split (SMAC 30 candidates / Song \& Gu 8-step subset / ISAC 30 intra-cluster searches,
each on 50 or 30 train instances; L2P-style 50 train instances × 200 evaluations; Hydra reuses the train-side configuration×instance matrix of the SMAC-style arm),
and the evaluation instances are used only for final execution and timing — no row of this table takes its value after searching on the evaluation instances.

Selection rule for the experimental domains: the comparison uses the three domains SC / CA / GISP (100 val instances per domain). The rule is objective — they are taken from the domains of the same batch
(see the batch discipline of §3.1) for which all opponent arms have been fully measured, rather than selected by results; among them CA is precisely the domain where the cross-domain transfer of this paper's method
fails (Appendix B.4), and it is deliberately retained. The scope of the comparison conclusions is limited to these three domains; the main experiments (Appendix B.4) cover 7 domains, and the two domain sets differ
(the resource-cost axis uses another 4 domains, see below).
Protocol boundaries. (i) Public implementations of prior methods are mostly incompatible with the solver stack used here (L2P-MIP \citealp{liu2024l2pmip} depends on ecole and on old versions of SCIP/PySCIPOpt),
and running them directly would split the solver protocol in two and make absolute times incomparable, so this paper reimplements their setups rather than their code, and the "category and citation" column of the table indicates the source;
(ii) budget alignment: SMAC-style search 30 configuration evaluations (= the 30 candidates of the domain-level search), the total budget of ISAC per-class search and of Song \& Gu subset selection
is likewise 30, and L2P-MIP-style per-instance search 200 evaluations/instance (above its scheduling lower bound of 153) additionally has a same-budget version (2 evaluations/instance);
(iii) tuning set = train, evaluation set = 100 val instances (the same batch of instances and the same denominator as the main-line experiments); all values are in solver-internal time (§3.1),
the numerator protocol of single arms and of the arms of this paper is given in the footnote, and the denominator = the internal clock of the replica of the 5 isomorphic default replicas in the same batch selected as the first to finish.
The experimental domains are SC / CA / GISP (the 300s preset); among them CA is the domain where the cross-domain transfer of this paper's method fails (Appendix B.4), and it is reported as well so as to avoid "picking domains by result".
Per-core resource readings and the accounting protocol are in Appendix B.8.

The domain set of the resource-cost axis is CFLP / MVC / SC (overlapping the comparison domains of Table \ref{tab:opponents-config}/\ref{tab:opponents-deploy} in the single domain SC): this axis needs only the two readings of content single-path and racing
and no configuration-class opponent arm, so it can cover more domains — the two domain sets differ out of protocol necessity, not out of inconsistent selection. Readings (solver-internal time, same protocol as Table \ref{tab:opponents-deploy}):
SC content single-path 0.9922 versus racing 0.7729; MVC content single-path 0.9902 versus racing 0.9882 (same batch, 100 keys, win rates 52\% versus 59\%);
CFLP content single-path 0.9253 versus racing 0.9166 (same batch, 100 keys, win rates 69\% versus 70\%);
the corresponding core-time ratios per core (5 × racing / content single-path) are SC 3.90 / MVC 4.99 / CFLP 4.95 (measured on the same batch for the three domains).
The content single-path reading takes the min of 5 replicas with the same content, while the racing reading takes the single reading of the winner in the pool, so this asymmetry pushes down the relative gain on the racing side, and we do not explain it in our favour.

\begin{table}[!htb]
\caption{Opponents at the deployment layer and the resource cost}
\label{tab:opponents-deploy}
\centering
\small
\tblscale{%
\begin{tabular}{>{\raggedright\arraybackslash}p{3.2cm}rrrp{0.9cm}>{\raggedright\arraybackslash}p{4.2cm}}
\hline
Deployment mode & SC & CA & GISP & cores & remarks \\
\hline
content single-path of this paper (no racing, single shot) & 0.992 & 1.181 & 0.921 & 1 & the content single-path form of the method of this paper \\
\(2K+1\) racing of this paper & 0.773 & 1.102 & 0.921 & 5 & deployment of this paper \\
\hline
core-time ratio per core (5 × racing / content single-path) & 3.90 & 4.66 & 5.00 & --- & same batch over the three domains (the content single-path reading is taken from the rebuilt wave of the content single-path and has been normalized to the same-batch denominator); the two cells CA / GISP of this row come from the comparison-domain batch (the same batch as Table \ref{tab:opponents-config}), and the domain set of the resource-cost axis (content single-path / racing) is CFLP / MVC / SC, whose core-time ratios CFLP 4.95 / MVC 4.99 / SC 3.90 are given in §A.3 \\
\hline
\end{tabular}}
\end{table}

Deployment-layer conclusions (the cost and the value of racing). (i) Racing trades 3.9--5.0× core-time for a solve-time gain, so it is a deployment means of "trading resources for time",
not free speedup; when a paper reports speedup it must report the resource cost as well. (ii) The value of racing is domain-specific: on SC it pushes content single-path 0.9922 down to 0.7729
(a significant gain, consistent with SC's 75\% np share in the racing pool); a same-batch re-measurement on MVC gives content single-path 0.9902 versus racing 0.9882
(a difference of 0.4\%, \(n = 100,\) within noise), so "almost zero gain" holds on that domain, but we do not claim a gain of racing on MVC; the same batch on CFLP gives
content single-path 0.9253 versus racing 0.9166 (a gain of 0.9\%, far smaller than the 4.95× core-time cost), so it belongs with MVC to the domains where "the cost buys no gain";
on this basis the gains of this paper decompose into two branches:
content quality and tail truncation (racing, see Proposition 1).

Reading status of prior work. (i) Reading scope: the domain-level configuration search and the method of this paper are the three domains SC / CA / GISP; per-instance configuration (L2P-MIP-style) /
SMAC / Song \& Gu / ISAC / Hydra-MIP, five categories in all, have been completed (three domains); the conclusions below are likewise limited to these three domains.
(ii) Of the 15 cells read for the domain-level configuration class, 12 fall in 0.969--0.998 (median 0.984), and only three cells on CA, namely SMAC-style (1.018), ISAC-style (1.005) and Hydra-MIP-style (1.131), are above the default, so within the 42-dimensional configuration space that is searchable here,
tuning the configuration to the domain-level optimum does not push end-to-end time below the default strategy,
and "tuning is enough" does not hold (the scope of this conclusion is given in Appendix B.12(e)). The same batch of readings also shows that per-instance configuration selection opponents (L2P-MIP-style 0.857 / 0.787 / 0.875,
SC / GISP / CA in order) are below the same-batch default strategy on all three domains, and on CA are also better than the racing arm of this paper (0.875 versus 1.102), while CA is precisely the domain where the cross-domain transfer of this paper fails —
the direction of this difference is unfavourable to this paper, which we do not explain in our favour, so "domain-level tuning is ineffective" and "the configuration space is ineffective" are not the same proposition;
the per-row details are in Table \ref{tab:opponents-config} (\S3.6) and in the per-row setup-reproduction details below. Read together with the corpus position of CA, this comparison only marks one boundary of applicability: CA is the domain with the richest corpus of this paper and yet a failed transfer (Appendix B.4), so the applicability of zero-shot sequence transfer does not expand monotonically with corpus richness, and we do not infer transfer success or failure from it.
(iii) Retrieval baseline: nearest-neighbour sequence retrieval in the same domain attains 0.9866 (65\%) on SC (same-batch denominator, \texttt{knn\_val\_v2}), weaker than 0.773 (93\%) of this paper, so learned content beats pure retrieval under the same protocol.
The discussion of the boundary between configuration space and action space is in Appendix B.12(e) — it gives the measured parameters of the change family of the solver emphasis preset
(13 / 16 items, excluding priority and trigger timing) and the determination that it is orthogonal to the action-space coordinates of this paper.

\subsection*{B.9 Paradigm Absorbability: Capabilities and Boundaries}
\label{sec:org9be1933}

Structure and boundaries. The decision object of this paper is the action sequence; the outputs of existing learning-based presolve schemes (configuration-vector tuning \citealp{liu2024l2pmip}, per-instance preset selection \citealp{li2025benloc},
subset/rule selection \citealp{song2024selection}; \citealp{xu2011hydramip}) can all be replayed as an action sequence (expanding the selected parameters, presets or subsets into concrete plugin-call sequences).
Hence for the framework of this paper the outputs of these works can in principle be absorbed as additional sequence corpus — this is a capability brought by "taking the sequence as the unified decision object", not an extra mechanism present anywhere;
the converse does not hold: decision objects such as configuration vectors and priorities can influence the call order only indirectly and cannot express the temporal dependencies among actions, so the sequence-type decisions examined in this paper are
not directly given by the decision object of existing methods (the same holds for the action surface newly added in this paper, such as probing-type propagators entering action-level decisions).
This asymmetry is a checkable methodological boundary and does not depend on any measured value.

Absorbability is not the same as profitability. The effect of absorption must be determined by experiment, and this paper does not presuppose its direction. Furthermore, if the corpus is instead constructed with per-instance hindsight optimal
labels, the base model in fact becomes worse — which also bounds the scope of the claims of this section. The quality and diversity of the corpus are more critical than
"optimal answers". Hence the claim of this section is "possesses the capacity for absorption", not "absorption yields improvement".

Paradigm absorbability (capabilities and boundaries): taking the sequence as the unified decision object makes the output of existing learning-based presolve uniformly replayable as an action sequence, so that it can in principle be absorbed as additional sequence corpus — this asymmetry is a checkable methodological boundary brought by "taking the sequence as the unified decision object" and does not depend on any measured value; the converse does not hold (configuration vectors and priorities can influence the call order only indirectly and cannot express the temporal dependencies among actions).
\subsection*{B.10 Related Work: Full Review and Supplements}
\label{sec:org509698d}

\subsubsection*{B.10.1 Reduction Rules and Round Structure of Presolve}
\label{sec:org8fd047f}

Achterberg et al. systematically summarized the reduction rules of MIP presolve and their effect in cutting model size and solving time\citeyearpar{achterberg2020presolve}; PaPILO implements presolve as an independent parallel library
that organizes 17 presolvers with "rounds + transaction conflict management"\citep{gleixner2023papilo}. The concept of the "round boundary" in this paper is of the same origin as that round structure, and the novelty of this paper does not lie in rounds themselves,
but in treating the round boundary as a generatable sequence structure — in the default strategy rounds are decided by a counter internal to the solver, whereas this paper allows the model to explicitly output round-boundary actions (the \(\bot_{\mathrm{round}}\) of Definition 1, §2.3).
\subsubsection*{B.10.2 Instance-Specific Configuration and Algorithm Selection}
\label{sec:orgceff735}

ISAC recommends algorithm configurations for different instance classes by clustering instance features\citep{kadioglu2010isac}, and its cost protocol includes a total-time decomposition \(T_{\mathrm{tot}} = T_{\mathrm{tune}} + T_{\mathrm{TTS}}\) that accounts for tuning time:
this paper follows that decomposition, classifying corpus collection and model training as offline \(T_{\mathrm{tune}}\) and the per-instance sequence generation and racing execution as online \(T_{\mathrm{TTS}}\) (§3.1 / §A.3).
Hydra-MIP searches configurations on subsets of instances and selects among them at runtime\citep{xu2011hydramip}; SMAC provides general model-based configuration search\citep{hutter2011smac}. The evaluation protocol of this family of work also carries lessons:
BenLOC points out that homogeneous datasets and permuted splits in it often cause data leakage and over-optimistic conclusions\citep{li2025benloc}, and it gives instance-level splits and a consistent-baseline specification — the evaluation of this paper adopts instance-level splitting.
\subsubsection*{B.10.3 Parallel and Racing Deployment}
\label{sec:org3e9b0cf}

The \(2K+1\) racing of this paper concurrently runs multiple candidate sequences at the presolve-sequence level and takes the one with the shortest solving time, additionally reporting the resource cost and the per-core efficiency
(the third layer of §3.6) — racing is a deployment means that trades core-time for solving time, not a free speedup.
\subsubsection*{B.10.4 MIP Presolve and Parameter Tuning}
\label{sec:org0e87e55}

MIP presolve is one of the core components of modern solvers: before branch and bound it performs structural reduction, variable reduction and bound tightening on the original problem, lowering the problem size and the difficulty of solving.
Classical techniques include constraint propagation, elimination of redundant constraints, detection of fixed variables, bound tightening and linear reasoning, among others\citep{achterberg2009scip}.
Traditional optimization relies mainly on heuristic rules and parameter configurations hand-designed by experts: solvers such as SCIP and Gurobi expose a large number of presolve parameters (priority, round limit, trigger timing, etc.),
with which users can adapt to a particular class of problems, but this requires deep domain knowledge and tuning experience. Instance-specific algorithm configuration is an automated tuning paradigm that has arisen in recent years:
the representative work ISAC recommends algorithm configurations for different instance classes by clustering instance features\citep{kadioglu2010isac};
L2P-MIP instead turns presolve optimization into fixed-dimensional configuration-vector learning, using a neural network to learn a mapping from instance features to presolve parameter configurations\citep{liu2024l2pmip}.
But such methods are all confined to the space of configuration parameters and cannot express scheduling decisions at the action level.

The three supplementary reviews — the reduction rules and round structure of presolve, instance-specific configuration and algorithm selection, and parallel and racing deployment — are in Appendix B.10.1--B.10.3 above.
\subsubsection*{B.10.5 Learning-Based Presolve: From Routine Selection to Parameter Configuration}
\label{sec:orge603131}

Song \& Gu learn the subset selection of "which presolve routines participate"\citeyearpar{song2024selection}: they decide only the participating set and not the call order, their decision object is the set of participating routines and does not cover the call order of actions;
L2P-MIP formalizes presolve optimization as a 42-dimensional configuration vector over the priority / timing / maxrounds
of 14 routines, searching near-optimal configurations per instance by simulated annealing and predicting them by supervised learning\citep{liu2024l2pmip}; this is the most directly comparable work to this paper and the principal opponent of the configuration space (see Table \ref{tab:opponents-config} of §3.6).
The relation between the action space of this paper and that one can be stated precisely as: the configuration space covers only the priority / timing / round parameters of each routine and cannot intervene directly in the call order of actions; this paper moves the decision object down from "static configuration" to "executable call sequence".
\subsubsection*{B.10.6 Independent Evidence for Order Sensitivity}
\label{sec:orgaba0711}

Learning-based methods have already succeeded in parts of MIP such as branching and cutting planes\citep{tang2020cutting,gasse2019exact}; order is equally important in these parts — the choice and order of cutting planes affect LP solving efficiency\citep{tang2020cutting}; CLCR reorders constraints by contrastive learning
and reports an average reduction of 30\% in solving time and an average reduction of 25\% in LP iterations\citep{zeng2025clcr}. These works show that "order is an underrated decision dimension in MIP solving";
this paper extends that observation to the atomic action sequence of the presolve stage and provides an executable injection and sequence execution facility (§2.4) that makes it amenable to controlled perturbation and measurement.
\subsubsection*{B.10.7 Combinatorial Solving and Algorithm Racing}
\label{sec:org3613500}

The core idea of portfolio solving is that the performance distribution of a content single-path policy is often heavy-tailed, with a few bad cases severely dragging down the whole, and that concurrently executing several candidates and taking the fastest can truncate this tail risk.
However, existing portfolio/racing methods mostly construct candidates at the solve stage or at the level of parameter configuration, and have not yet gone down to the level of atomic action sequences in presolve;
this paper shows that sequence differences at the presolve stage have a significant effect on subsequent solving performance, and that introducing a racing mechanism at the presolve level has independent value.

Positioning of existing work. On the MIP side, learning-based presolve is carried out entirely within the configuration space: L2P-MIP searches over the
priority / timing / round of 14 routines, a 42-dimensional configuration vector in all (ICLR 2024), and Song \& Gu learn the selection of routine combinations (ICMLC 2024).
The configuration space describes only the global on/off switches, priorities and round caps of actions, and does not take the cross-round call order itself as the decision object.
The reason such a formulation has not been established, apart from the modelling viewpoint, also lies in an enabling prerequisite: solvers do not provide a sequence execution interface.

Existing work can be divided into three classes: manual heuristic tuning, instance configuration tuning, and algorithm-portfolio racing. Mainstream solvers such as SCIP and Gurobi rely on expert experience to preset fixed priorities, timings and maximum numbers of rounds,
reusing the same scheduling rules for all instances, which makes it hard to adapt to diverse instance structures\citep{bixby2007progress,achterberg2009scip};
the representative of instance configuration tuning, L2P-MIP, turns presolve optimization into fixed-dimensional configuration-vector learning, searching the priority, timing and round parameters of presolve routines by simulated annealing
and completing the mapping from instances to configurations with a neural network, but it is confined to the space of configuration parameters, can only control the order of action execution indirectly, cannot express complex scheduling behaviour such as the reversal of action order across rounds,
and cannot intervene in the scheduling decisions of plugins such as constraint handlers and propagators\citep{liu2024l2pmip};
work on algorithm portfolios / per-class configuration such as ISAC and work on per-instance configuration such as BenLOC\citep{kadioglu2010isac,li2025benloc} achieve speedups by taking the best result over several parameter configurations,
but they construct candidates only within the parameter-configuration space, which is bounded by the parameter knobs exposed by the solver. None of the three classes of work has touched the dimension of the call order and round structure of presolve actions;
moreover, the overwhelming majority of learning-based schemes depend on intermediate-state feedback exposed by the solver, whereas general black-box MIP solvers do not support feeding a custom presolve action sequence directly,
and there is no execution substrate on which to study presolve order effects in a controllable way.
\subsection*{B.11 Details of the Method and Implementation}
\label{sec:orgb0d7c2f}

\subsubsection*{B.11.1 Model Architecture and Training Objective}
\label{sec:org3767199}

Presolve planning and language generation are structurally isomorphic: their input is the initial description of the problem, their output is a variable-length sequence of discrete action types, and the actions exhibit complex contextual dependencies among themselves. Accordingly,
this paper explicitly reformulates presolve planning as a language sequence generation task---problem features as input, the presolve action sequence as output---and expresses sequence generation itself as the standard language modelling objective of "predicting the next action given the preceding context".
Formally, let \(x\) be an MIP instance with numerical feature vector \(u\in\mathbb{R}^d\) (the model input), and let \({\mathcal A}=\{1,\ldots K\}\) be the finite dictionary of presolve atomic actions. At each decoding step the model outputs an action \(a_t\in {\mathcal A}\) ,
thereby expressing the generation distribution as the standard autoregressive conditional probability \(p_\theta(a_t|a_{0,\ldots,t-1}, x)\) .

Based on this modelling paradigm, this paper adopts a Transformer decoder-only architecture as the backbone. Its causal attention mechanism allows the action generation at each step to depend directly on the complete order-preserving historical context, and no longer on the information bottleneck brought by step-by-step state summarization;
its absolute positional encoding naturally preserves the temporal topological structure among actions, eliminating the problem of "order blindness" at the root.

This framework requires only a single initial-state input, and the subsequent generation process is based entirely on the model's internal recurrence, with no need to access solver-internal state during generation---which, under the realistic constraint that MIP solvers do not expose an intermediate-state interface,
makes it the only engineering-feasible sequential decision-making scheme.

Presolve execution trajectories of the SCIP solver on the training-set MIP instances are collected through a custom logging interface. Through quantitative indicators such as the number of fixed variables, the reduction in non-zeros, the number of deleted constraints and the magnitude of bound tightening, this interface judges whether an action produces a substantive reduction effect,
and records it only when the action takes effect. The training objective is the cross-entropy loss of standard autoregressive language modelling: \({\mathcal L}_{\mathrm{SFT}}(\theta)=-\sum\limits_{t=1}^T \log p_\theta(a_t|a_{0,\ldots,t-1},x)\)
This stage enables the model to acquire the basic "grammar" of presolve action sequences, providing an initialization for the subsequent improvement of sequence quality.
It should be pointed out that the above screening mechanism creates a structural discrepancy between the "effective action sequences" recorded in the log and the complete action sequence actually executed inside SCIP (which contains a large number of ineffective attempts); although the model approaches the distribution of SCIP's effective reduction trajectories,
it is not an exact simulation of its default policy, its fidelity is subject to an inherent limitation, and the SGM can directly reflect the degree of loss in sequence quality.
\subsubsection*{B.11.2 Execution Interface for Presolve Sequences}
\label{sec:org7c802aa}

Research setup (trace-only). Only two kinds of data are observable: an instance's initial features \(x\) and the action sequence derived from the execution trace of the default strategy;
the intermediate states of presolve are not observable.

Conducting research under the formalism of Definition 1 first requires answering a methodological question: in a black-box
solver whose internal state is unobservable and whose order assumptions are embedded in the implementation, how can the semantic faithfulness of "executing a given action sequence" be guaranteed? If the execution semantics are inconsistent with the solver's native behaviour,
all subsequent measurements and conclusions lose their meaning. This paper gives two mechanisms:

(a) Identification of effect-equivalent actions. The raw call log contains a large number of routine calls that are skipped or have no actual effect, and taking it directly as a sequence would introduce spurious actions
(diluting the supervision signal and corrupting the sequence semantics). This paper determines whether an action is effective by whether an observable structural variable has changed,
thereby projecting the call log into a semantic action sequence. This criterion is an implementation-level signal, and the monitoring is potentially lossy: if there exists a structural change whose variation is not
captured by the current monitored variables, the corresponding action is judged ineffective and discarded---this paper does not claim monitoring completeness (see Appendix B.12)
(b) Sequence injection along the default execution path. The internal state and order assumptions of the presolve phase make it impossible to insert an external sequence directly.
Our approach is to make the solver accept an externally specified action sequence while preserving its default execution semantics,
so that what is executed is the solver's real presolve behaviour rather than a reimplementation of the presolve logic.
Fidelity is the core metric of this mechanism and the precondition for the comparability of all subsequent conclusions.

Stock SCIP provides no interface for executing a given presolve action sequence (its presolver plug-in interface exposes only priority and rounds), so this paper modifies the SCIP 10.0.1 source: recording and controlled-execution hooks are added at the three plug-in execution entry points \texttt{SCIPpresolExec} / \texttt{SCIPconshdlrPresolve} / \texttt{SCIPpropExec}, and six new C interfaces are exported and exposed through the PySCIPOpt binding. The unmodified official build serves as the control arm (Appendix B.1).

License and attribution. The solver used here is SCIP 10.0.1 (Apache License 2.0; \url{https://github.com/scipopt/scip}) together with SoPlex 8.0.x (Apache License 2.0);
the modifications above are confined to the three plug-in execution entry files (src/scip/presol.c, src/scip/cons.c, src/scip/prop.c) and to the PySCIPOpt binding (MIT licensed, also modified).
The modified files retain their upstream copyright notices and carry a notice of modification, and a per-file inventory of the changes will be provided with the follow-up public release; this work is an independent derivative and is not affiliated with, endorsed by, or supported by ZIB or the SCIP project.

(c) Skip semantics for plugins not listed. In replay mode, plugin calls that do not appear in the sequence are skipped, and only the actions that once took effect are executed, each exactly once,
so that the multi-round sweeps and ineffective attempts of the default presolve are no longer repeated---for a costly component such as probing,
none of its attempts that produced no effect are replayed. Fidelity thereby acquires a testable statement: the reduction set produced by a replayed sequence
should be identical to that at recording time (its test is in Appendix B.1).

The action space is extended from a single plugin family (routines) to the unified action space of Definition 1 (three families: presolve routines / propagators / constraint handlers; the propagator family is not triggered in the corpus of this paper),
bringing two classes of actions with different mechanisms, "structural reduction" and "domain propagation", into the same decision domain that is comparable, rankable and composable; round boundaries and the execution contract (Definition 2)
are simultaneously included in the generation object, so "in what order, at which round, and with which delivery mode to execute" are unified into choices on one and the same sequence (their formal status is given in Remark 1 of this section). This extension has no precedent in existing learned presolve work:
L2P-MIP is a 42-dimensional configuration vector over 14 presolvers (a single family)\citep{liu2024l2pmip}; propagators and constraint handlers are MIP-specific plugin families and are not within that configuration space.

The first concrete consequence of this extension: the presolve callback of the constraint handler enters action-level decisions---in our measurements it accounts for
 85-100\% of all executed actions, whereas parameter-level methods 
 have no parameters of that plugin family to tune, and its call order is likewise not determined by the configuration.

The second concrete consequence of this extension: one of the most expensive components of the presolve phase (probing) has no corresponding parameter to tune in existing parameter-level methods---in SCIP, probing registers as a propagator
and participates in presolve (since SCIP 3.1) rather than as an independent presolve routine, so the 42-dimensional space of L2P-MIP (the priority / timing / round of 14 presolvers\citealp{liu2024l2pmip}) cannot reach it;
at the interface layer of this paper, "whether to execute probing, at which round to execute it, and at which timing stage to execute it" is a per-instance action-level decision (the actions of that plugin family are not triggered in the corpus of this paper, see above).

A checkable gap at the formal level: the control of the configuration space over the call order is indirect---the configuration only gives the priority / timing / rounds of each routine, and the solver decides on that basis which call sequence is actually executed, with no direct intervention in the order of constraint handler callbacks;
whereas this paper takes the order itself as the object of generation. The shuffling experiment (§3.3; Appendix B.3) shows that different orders of the same multiset can produce significantly different end-to-end runtimes.
\subsection*{B.12 Applicability Boundaries of the Method}
\label{sec:orgca44094}

The conclusions of the method in this paper are based on the single solver version SCIP 10.0.1, and the action space is constrained by the callback interfaces the solver exposes. Results at the instance level are perturbed by the solver's random seed; all conclusions are statistical and we do not promise that every instance will get better.

Furthermore, in the current experiments the number of triggers of propagator actions in the corpus is 0; this capability of the unified action space is at present only formally available, and its actual gain remains to be further verified.

The testable consequences of the above formal conclusions are verified respectively in the necessity of order (Appendix B.3), the availability of the gain (§3.4) and the transfer criterion (Appendix B.4).
\subsubsection*{B.12.1 Boundaries at the Solver and Action-Space Level}
\label{sec:orgb1201}

Boundaries: (a) the action space is limited by the set of callbacks exposed by the modified solver and does not cover all intervention points inside the solver;
(b) "round" as a concept is not introduced by this paper (the round-based presolve loop of PaPILO \citealp{gleixner2023papilo}; the configuration of L2P-MIP likewise contains a per-presolver round scalar \citealp{liu2024l2pmip}),
the novelty of this paper lies in rounds as a generatable sequence structure; (c) the opponents at every layer are reimplementations of setups within the same arena, and numbers are not juxtaposed with prior work whose reporting protocol differs;
(d) the determination of effective actions is based on implementation-level signals of internal structural changes of the solver, with a potential risk of lossy screening.
\subsubsection*{B.12.2 Relation to the Configuration Space}
\label{sec:orgb1202}

(e) Boundary of the scope of the configuration space: the configuration-space comparison of this paper is limited to the priority / trigger timing / round of presolve routines (42 dimensions) and does not include the solver's solution-emphasis presets; therefore the conclusion of this paper that "configuration is insufficient to replace sequence decisions" holds only within that configuration space.
This emphasis consists of two classes of factors---(i) more rounds and restarts: round boundaries are within the action space of this paper (the NEW\_ROUND action), whereas the base corpus is taken from the execution trajectories of the default policy,
whose round structure is nearly unique (Appendix B.3), so the base corpus does not contain the direction of "adding rounds"; (ii) stronger reduction strength: we measure that this preset (the solver's built-in emphasis)
changes only 13 parameters (aggressive) / 16 (fast), concentrated in the per-routine round limits (\texttt{presolving/*/maxrounds} from 0 to -1), the restart threshold
(\texttt{presolving/restartfac} from 0.025 to 0.0125) and a switch (\texttt{constraints/setppc/cliquelifting} changed from False to True). These parameters contain no priority /
trigger-timing item, so they are orthogonal to the coordinates of this paper's action space (routine × trigger timing × round) and are not generatable sequence actions.
The claim of this paper therefore narrows to: under the round structure of the default policy, sequence-level decisions can significantly change end-to-end time; round boundaries are themselves generatable actions of this space, but the corpus contains no example of a round structure beyond the default policy's, so whether adding rounds can yield gains is not measured in this paper.
\subsubsection*{B.12.3 Evidence Granularity and Declarations}
\label{sec:orgb1203}

(f) Metric granularity: this paper attributes differences in the search phase only qualitatively and does not report the number of branching nodes or the number of LP iterations (collecting these two would align directly with the metrics commonly used by L2P-MIP \citealp{liu2024l2pmip}
and by the order-sensitivity literature \citealp{tang2020cutting,zeng2025clcr}, and is left to future work).

Boundaries of the characterization:
① The characterization is for a single solver and its version (SCIP), and the conclusions are not extrapolated to other solvers;
② the corpus consists of execution trajectories of the default policy and characterizes the behaviour distribution of the default policy, not some optimal sequence distribution;
③ \(N_{\mathrm{eff}}\) takes the entire sequence as the atomic unit and does not characterize action-level local entropy;

Declarations of scope:
(1) Online competition (racing / portfolio) is an existing MIP technique, and this contribution does not lie in the novelty of the mechanism,
but in the sequence-level decomposition of the sources of gain and the characterization of transfer conditions under this new problem formulation;
(2) the racing pool does not contain a default-policy path, so SGM<1 is a pure selection gain rather than a structural guarantee (random content 1.16-1.20 is the counterexample);
(3) mechanisms A/B on known domains can in principle be reproduced by "directly executing a sequence known to be effective on that domain"---the value of the model is concentrated in mechanism C;
(4) the cross-structural evidence of zero-shot is based on one unseen domain (NNV); all 6 LOO test points have been completed (CA/CFLP/GISP/MIS/MVC/SC each removed once);
(5) the characterization of transfer applicability is outside the scope of this paper;
(6) a single solver and its version; the multi-seed recheck of racing and of the isomorphic denominator (independent solver seeds) is given in Appendix A.2, item (5), which covers six domains × three solver seeds and shows that the seed-to-seed fluctuation does not change the per-domain speedup/tie verdicts;
single-instance runtime is sensitive to the seed (the per-domain sensitivity readings are likewise given in Appendix A.2, item (5)), so instance-level conclusions are always given in statistical form; per-class runtime is confounded by instance difficulty, and only controlled losses can be read causally;
 (7) the magnitude of mechanism A is jointly delimited by a three-configuration control (default / replay / presolve fully disabled (np), the three arms in Appendix B.7) and by the "consistency of the replayed reduction set" test;
 the variant that "disables only probing while keeping the remaining presolve components" is outside the scope of our measurements and, like the former, is left to future work; the claims of this paper about mechanism A are limited to its existence and its direction.
\subsubsection*{B.12.4 Boundary Observations on Cross-Domain Transfer}
\label{sec:orgfce51ec}

 The asymmetry of transfer success: the zero-shot transfer of CA (the richest corpus) fails (1.1121), whereas that of GISP (the most degenerate corpus) succeeds (0.9357);
the characterization of the applicability range is outside the scope of this paper, and the per-point readings are given in Appendix B.4.

No validated predictor of transfer success is available: the quantities examined---the share of total time spent in presolve, corpus richness, and domain-semantic similarity---do not separate the seven points, since the failure occurs on the domain with the richest corpus (CA) while the strongest gains occur on the most degenerate corpus (GISP) and on the most dissimilar domain (NNV). We report this as a boundary of the method rather than as a selection rule; constructing and validating such a rule would require more transfer points than the seven available here.

The \(2K+1\)-way concurrent racing mechanism requires multi-core compute. Although the inference overhead of sequence generation is small, the overall system requires multi-core parallel resources. In practical deployment, the concurrency \(K\) and the expected gain must be weighed against the available compute resources.
This cost deserves to be quantified: the core-time ratio per core of racing relative to content single-path is 3.9--5.0 times,
and on MVC and CFLP this ratio consists almost entirely of "pure cost" (across the two domains racing yields virtually no time gain over content single-path: differences on the same batch of 0.4\% and 0.9\%)
Thus the justification for racing depends on whether the target domain exhibits tail risk amenable to truncation (in this work: present for SC, absent for MVC / CFLP).
Under the ISAC cost decomposition \citep{kadioglu2010isac}, corpus collection and training in this work belong to the offline \(T_{\mathrm{tune}}\)
(one-off and reusable across domains), whereas per-instance generation and racing belong to the online \(T_{\mathrm{TTS}}\); the two must be reported separately, otherwise "5 times more expensive" and "2--3 orders of magnitude cheaper" would mask each other (§A.3).
\section*{Appendix C Notation and Symbols}
\label{sec:orgnotc}

This appendix collects the meaning of the mathematical notation, formulas and abbreviations used throughout the paper; the main text refers to this appendix instead of repeating the definitions.

\clearpage
\subsection*{C.1 Mathematical notation}
\label{sec:orgnotc1}

\captionof{table}{Mathematical notation used in this paper}
\label{tab:notation}

\begin{center}
\small
\begin{tabular}{@{}p{0.19\textwidth}p{0.75\textwidth}@{}}
\hline
Symbol & Meaning (where it appears) \\
\hline
\(x\) & an MIP instance (objective and constraint data) (§2.2) \\
\(\theta\) & trainable parameters of the sequence-generation model (§2.2) \\
\(a_t\) & the atomic action executed at step \(t\) of an action sequence (§2.2) \\
\(a_{0,\ldots,t-1}\) & the prefix already executed before step \(t\), \((a_0,\dots,a_{t-1})\) (§2.2) \\
\(a_0\) & the initial-state token (instance features with the start token), included in the prefix and never predicted (§2.2) \\
\(\bot_{\mathrm{round}}\) / \(\bot_{\mathrm{eos}}\) & the round-boundary and the end-of-sequence markers in a generated sequence (\S2.3) \\
\(T\) & completion time (solver-internal clock); in sequence contexts also the index of the last token, with token sequence \(a_0,\dots,a_T\) and predicted actions \(a_1,\dots,a_T\) (§2.4 / §2.2) \\
\(T_i\) & completion time of the \(i\)-th candidate; racing takes \(\min_{i\le K}T_i\) (§A.1) \\
\(\mathcal{A}\) & the unified atomic action space; its three subsets \(\mathcal{A}_{\mathrm{presol}}\) / \(\mathcal{A}_{\mathrm{prop}}\) / \(\mathcal{A}_{\mathrm{cons}}\) are the actions produced by standalone presolve routines, propagator callbacks and constraint-handler callbacks (§2.3) \\
\(r\) / \(\mathcal{R}\) & one candidate sequence in the pool / the set of candidate sequences (the pool) (§2.4) \\
\(K\) & number of candidates per content arm and per fallback arm (§2.5) \\
\(2K+1\) & pool size \(=1\) np arm \(+K\) custom arms \(+K\) fallback arms (§2.5) \\
\(N\) & number of isomorphic default replicas on the denominator side; the main protocol matches the pool, \(N=2K+1\) (§B.2.2) \\
\(n\) & number of evaluation instances (keys): 100 per domain, 550 for NNV (§3.1 / §3.5) \\
window & number of replicas launched per arm at the same time; about 60 cores in total (§3.1 / §B.2.2) \\
\(t_i\) & runtime of the \(i\)-th instance (the per-instance reading in SGM) (§3.1) \\
\(Q_p(\cdot)\) & the upper \(p\)-quantile function of a distribution (§A.1) \\
\(p\) & quantile level (\(p=0.90\) written \(p_{90}\)); in \(p_\theta\) it denotes the policy distribution (§3.1 / §2.2) \\
\(\alpha\) & upper-tail index of the runtime distribution (Pareto tail index; smaller means heavier tail) (§A.1) \\
\(\overline F(t)\) / \(L(t)\) & the upper-tail function and its slowly varying factor: \(\Pr(T>t)=L(t)\,t^{-\alpha}\) (§A.1) \\
\(\Delta t\) & runtime difference of the same instance between two paired measurements; seed sensitivity uses \(|\Delta t|\div\min(t_i,t_j)\) (§A.2) \\
\(\Delta\) & paired difference between two arms' readings; the \(\Delta\) ratio column is shuffled minus ordered (§B.3) \\
\(\mathrm{SGM}(t)\) & main metric \(=\bigl(\prod_i (t_i+1)\bigr)^{1/n}-1\) (shift \(=1\) s, timeouts counted at that instance's limit); a ratio of two arms' SGM below \(1\) means speedup over the default policy (§3.1) \\
\(N_{\mathrm{eff}}\) & effective number of sequences \(=\) the number of distinct presolve sequences used by different instances within a domain; the superscripts seq / mult denote the order-preserving and order-ignored (multiset) counts (§B.3) \\
\(\rho_{\mathrm{order}}\) & order-invariance degree \(=N_{\mathrm{eff}}^{\mathrm{seq}}/N_{\mathrm{eff}}^{\mathrm{mult}}\) (§B.3, Definition 5) \\
\(T_{\mathrm{tune}}\) / \(T_{\mathrm{TTS}}\) / \(T_{\mathrm{tot}}\) & offline tuning cost / online solving time / total time \(T_{\mathrm{tot}}=T_{\mathrm{tune}}+T_{\mathrm{TTS}}\) (§A.3) \\
\(Q, K\) & query and key projections in the decoder's self-attention, carrying rotary position embeddings (RoPE) \\
\(f, g\) & two presolve atomic actions, used only to illustrate non-commutativity in Property 1 and in the expressiveness argument of Proposition 2 \\
\(E_c, E_0, E_D\), \(c\) & the execution mapping, the two contract variants and the contract itself (Definition 2) \\
\hline
\end{tabular}
\end{center}

\subsection*{C.2 Explanation of the formulas}
\label{sec:orgnotc2}

\begin{itemize}
\item Sequence-generation model \(p_\theta(a_t \mid a_{0,\ldots,t-1}, x)\) (§2.2): actions are generated token by token on instance \(x\), and step \(t\) is conditioned on the prefix \(a_{0,\ldots,t-1}\) already generated; the probability of a whole sequence is the product of the stepwise conditional probabilities. This is the formal landing point of recasting how presolve proceeds as a probability distribution over action sequences.
\item Unified atomic action space \(\mathcal{A}=\mathcal{A}_{\mathrm{presol}}\cup\mathcal{A}_{\mathrm{prop}}\cup\mathcal{A}_{\mathrm{cons}}\) (§2.3): the action space is the union of three families; the facility can intercept and execute all three at the interface level, while the corpus and the actual replay path of this paper involve two of them.
\item Tail truncation of the racing gain \(\frac{Q_p(\min_{i\le K}T_i)}{Q_p(T)}\sim(1-p)^{(1-1/K)/\alpha}\) (§A.1, Proposition 1): when \(K\) candidates race and the first to finish wins, its \(p\)-quantile runtime equals the \(1-(1-p)^{1/K}\) quantile of a single candidate; the ratio depends only on the quantile level \(p\), the number of candidates \(K\) and the upper-tail index \(\alpha\), and decreases monotonically as \(K\) grows, \(\alpha\) shrinks or \(p\) rises---racing is thus a tail truncator rather than a mean-quality booster. Numerical checks and the optimality boundary on \(K\) are given in §A.1.
\item Main metric \(\mathrm{SGM}(t)=\bigl(\prod_i (t_i+1)\bigr)^{1/n}-1\) (§3.1): the shifted (\(+1\) s) geometric mean of the per-instance runtimes minus one; the paper reports the ratio of two arms' SGM, where a value below \(1\) means speedup over the default policy.
\item Order-invariance degree \(\rho_{\mathrm{order}}=N_{\mathrm{eff}}^{\mathrm{seq}}/N_{\mathrm{eff}}^{\mathrm{mult}}\) (§B.3, Definition 5): the ratio of the order-preserving to the order-ignored (multiset) count, measuring how strongly the action order affects the final reduction result.
\end{itemize}

\subsection*{C.3 Abbreviations and names}
\label{sec:orgnotc3}

\begin{itemize}
\item ORDO: the framework of this paper---Operation-level Round-aware Dynamic Ordering for MIP Presolve, after the Latin \emph{ordo}; see §2.
\item SGM: the main metric (speedup reading), see item 4 of C.2 (§3.1).
\item TTS: online solving time (time-to-solution), which with the offline \(T_{\mathrm{tune}}\) forms the ISAC decomposition (§A.3).
\item LOO: leave-one-domain-out construction (§3.5).
\item B\&B: the branch-and-bound solving stage (see Figure \ref{fig:pipeline}).
\item EOS: end-of-sequence token; both generation and evaluation truncate at the first EOS per instance (§3.1).
\item np / custom / fallback: the three kinds of arms in the pool---presolve fully disabled / execute only the model's sequence / the sequence followed by the default presolve (§2.5).
\item default: the solver's default presolve policy, which also constitutes the isomorphic-load denominator (§2.5).
\item baseA / k3c: corpus variant names, defined in Appendix B.5.
\item Domain abbreviations (CA / SC / MIS / MVC / GISP / CFLP / NNV / MMCN / LB and others): see the dataset description in §3.1.
\end{itemize}

\end{document}